\documentclass[lettersize,journal]{IEEEtran}
\usepackage{amsmath,amsfonts}
\usepackage{algorithmic}
\usepackage{array}
\usepackage[caption=false,font=normalsize,labelfont=sf,textfont=sf]{subfig}
\usepackage{textcomp}
\usepackage{stfloats}
\usepackage{url}
\usepackage{verbatim}
\usepackage{graphicx}
\def\BibTeX{{\rm B\kern-.05em{\sc i\kern-.025em b}\kern-.08em
    T\kern-.1667em\lower.7ex\hbox{E}\kern-.125emX}}
\usepackage{balance}

\usepackage{tabularx}
\usepackage{hyperref}
\usepackage{booktabs}
\hypersetup{
	hypertex=true,
	colorlinks=true,
	linkcolor=blue,
	anchorcolor=blue,
	citecolor=blue
}
\usepackage{multirow}
\usepackage{booktabs}  
\usepackage{cellspace}  
\usepackage{hyperref}

\begin{document}
\title{Cross-Modal Pre-Aligned Method with Global and Local Information for Remote-Sensing Image and Text Retrieval}

\author{\IEEEauthorblockN{Zengbao Sun, Ming Zhao, Gaorui Liu, Andr\'{e} Kaup, \emph{Fellow, IEEE}}
	
\thanks{This work was supported in part by the National Natural Science Foundation of China under Grant 62271302, Grant 62101316; in part by the Shanghai Municipal Natural Science Foundation under Grant 20ZR1423500. (Corresponding authors: M. Zhao, G. Liu)}
\thanks{Z. Sun and M. Zhao are with the Department of Information Engineering, Shanghai Maritime University, Shanghai 201306, China. (e-mail: sunzb98@163.com; zm\_cynthia@163.com).}

\thanks{G. Liu is with Key Laboratory of Intelligent Infrared Perception, Shanghai Institute of Technical Physics, Chinese Academy of Sciences, Shanghai 200083, China (e-mail: liugaorui@sitp.ac.cn).}

\thanks{A. Kaup is with the Chair of Multimedia Communications and Signal Processing, Friedrich-Alexander University Erlangen-N$\ddot{u}$rnberg,
	Cauerstr. 7, 91058 Erlangen, Germany. }
\thanks{Manuscript received April 19, 2021; revised August 16, 2021.}
}

\markboth{Journal of \LaTeX\ Class Files,~Vol.~14, No.~8, August~2021}%
{Shell \MakeLowercase{\textit{et al.}}: A Sample Article Using IEEEtran.cls for IEEE Journals}


\maketitle

\begin{abstract}
In recent years, remote sensing cross-modal text-image retrieval (RSCTIR) has attracted considerable attention owing to its convenience and information mining capabilities. 
However, two significant challenges persist: effectively integrating global and local information during feature extraction due to substantial variations in remote sensing imagery, and the failure of existing methods to adequately consider feature pre-alignment prior to modal fusion, resulting in complex modal interactions that adversely impact retrieval accuracy and efficiency. 
To address these challenges, we propose a cross-modal pre-aligned method with global and local information (CMPAGL) for remote sensing imagery. Specifically, we design the Gswin transformer block, which introduces a global information window on top of the local window attention mechanism, synergistically combining local window self-attention and global-local window cross-attention to effectively capture multi-scale features of remote sensing images. Additionally, our approach incorporates a pre-alignment mechanism to mitigate the training difficulty of modal fusion, thereby enhancing retrieval accuracy. Moreover, we propose a similarity matrix reweighting (SMR) reranking algorithm to deeply exploit information from the similarity matrix during retrieval process. This algorithm combines forward and backward ranking, extreme difference ratio, and other factors to reweight the similarity matrix, thereby further enhancing retrieval accuracy. Finally, we optimize the triplet loss function by introducing an intra-class distance term for matched image-text pairs, not only focusing on the relative distance between matched and unmatched pairs but also minimizing the distance within matched pairs. Experiments on four public remote sensing text-image datasets, including RSICD, RSITMD, UCM-Captions, and Sydney-Captions, demonstrate the effectiveness of our proposed method, achieving improvements over state-of-the-art methods, such as a 2.28\% increase in mean Recall (mR) on the RSITMD dataset and a significant 4.65\% improvement in R@1.
The code is available from \emph{https://github.com/ZbaoSun/CMPAGL}.
\end{abstract}


\section{Introduction}
\IEEEPARstart{W}{ith} the rapid advancement of Earth observation technologies, remotely sensed imagery data has burgeoned, proffering unprecedented opportunities to explore and comprehend our planet \cite{ma2015remote}, \cite{chi2016big}. 
These data not only encompass a wide gamut of geographic, climatic, and ecological information \cite{srivastava2019understanding}, \cite{borana2019hyperspectral}, \cite{amani2020google}, but also exist in various forms and resolutions, proffering invaluable insights for environmental monitoring, resource management, urban planning, and numerous other domains. 
However, effectively managing and retrieving pertinent information from these massive datasets \cite{cheng2022nwpu} remains a significant quandary. The emergence of remote sensing cross-modal text-image retrieval (RSCTIR) techniques has provided a new perspective for addressing this issue. 
By enabling users to query images using natural language descriptions or retrieve textual information pertinent to a given remote sensing image, RSCTIR bridges the chasm between human intuitive expression and complex remote sensing data. 
This technology \cite{abdullah2020textrs}, \cite{cheng2021deep}, \cite{lv2021fusion} not only facilitates more direct and convenient information retrieval but also contributes to unveiling implicit relationships within the data, thereby fostering a deeper understanding of the intricate earth systems. 
Nevertheless, accurately bridging the semantic chasm between these heterogeneous visual and textual modalities remains a formidable task due to their inherent modal discrepancies. 
Visual data is typically high-dimensional, encompassing rich spatial information \cite{sippel2023cross}, \cite{grosche2023image}, while textual data is one-dimensional and contains abstract semantic information. 
These two modalities evince fundamental discrepancies in their modes of expression and information density, rendering cross-modal retrieval a complex quandary.

Currently, the crux of mainstream RSCTIR methods can be bifurcated into two main components: feature extraction for images and text, and modal fusion between images and text. 
Predicated on the complexity of feature extraction and modal fusion, mainstream methods can be categorized into two principal classes:

The first class consists of dual-branch networks \cite{yuan2022exploring}, \cite{yuan2022remote}, \cite{yuan2021lightweight}, \cite{zhang2023exploring}, \cite{zhang2023hypersphere}, which employ complex network structures to separately perform deep feature extraction for images and text, followed by relatively simple modal fusion approaches. 
Typically, the image branch employs  convolutional neural networks or vision transformers \cite{dosovitskiy2020image} to capture visual features, while the text branch utilizes models such as LSTM \cite{sak2014long}, GRU \cite{chung2014empirical}, or text transformers \cite{vaswani2017attention} to extract textual features. 
Ultimately, the features from both modalities are mapped into a common embedding space, and their similarity is calculated to obtain retrieval scores.
The common characteristic of these dual-branch network methods is their dedication to fully leveraging the unique structure of each modality, striving to extract deep semantic features from both visual and textual modalities, but adopting relatively simple operations for modal fusion. 
Some representative works include the asymmetric multi-modal feature matching network (AMFMN) proposed by Yuan et al. \cite{yuan2022exploring}, which designed an MVSA module to extract salient visual features from remote sensing images and utilized a bidirectional GRU to capture textual semantics, thereby enabling salient visual features to guide the extraction of textual features. 
Another example is the global and local information-based retrieval (GaLR) framework by Yuan et al. \cite{yuan2022remote}, where a multi-level information dynamic fusion (MIDF) module is employed for image feature extraction, while a bidirectional GRU is utilized for textual feature extraction. Chen et al. \cite{chen2023multiscale} introduced the multi-scale salient image-guided text alignment (MSITA) method, aiming to learn salient visual information while also adopting a strategy of complex feature extraction for images and text separately, followed by simple feature fusion.

The second class comprises single-branch networks, which perform only simple feature embedding for images and text, and then feed these embedded features as input into a single transformer model for complex modal fusion. 
The goal of this model is to learn a unified cross-modal representation, thereby bridging the semantic chasm between visual and textual modalities.
One representative work in this category is the interactively enhanced feature transformer (IEFT) model proposed by Tang et al. \cite{tang2023interacting}. This model takes remote sensing images and text as a whole input and learns the inherent associations between images and text through alternating self-attention and cross-attention mechanisms. 
This approach aims to initially embed images and text as basic features and then perform thorough modal interaction and fusion within a complex transformer model, thereby further mitigating the semantic discrepancy between modalities.

While existing dual-branch and single-branch RSCTIR methods have made progress, several quandaries remain to be addressed. Dual-branch networks focus on utilizing the unique structure of each modality to extract deep features but may struggle to effectively model cross-modal relationships during modal fusion. Conversely, single-branch networks emphasize capturing cross-modal information but may sacrifice the expression of individual modal details.

To address these quandaries, inspired by the align before fuse (ALBEF) approach \cite{li2021align}, we propose a cross-modal pre-alignment method predicated on global and local information from remote sensing images. This method comprises an image encoder, a text encoder, and a multi-modal encoder. 
Before inputting the extracted image and text features into the multi-modal encoder for fusion, we perform pre-alignment, mitigating the training complexity of modal fusion by aligning visual and textual features prior to integration, thereby alleviating initial modal discrepancies and fostering more accurate cross-modal matching during the retrieval process.

Furthermore, existing methods grapple with capturing the complex details and multi-scale information present in remote sensing images \cite{yuan2022remote}, \cite{chen2023deep}, \cite{zhao2023multitask}.
Inspired by the Swin transformer \cite{liu2021swin} and GCViT \cite{hatamizadeh2023global}, we propose an image encoder predicated on global and local window attention and design the global-swin (Gswin) transformer block. 
The Gswin, through its upper and lower branches, performs attention with different contents of local windows and the same global window. 
This synergistically integrates the global and local information of remote sensing images, thereby facilitating the capture of deep-level features more efficaciously.

Moreover, existing methods either disregard the bidirectional ranking information in the original similarity matrix or fail to fully utilize the information in the original similarity matrix for reranking \cite{yuan2022remote}, \cite{wang2019matching}. Consequently, we propose a similarity matrix reweighting (SMR) reranking algorithm that fully exploits the information contained in the original similarity matrix. 
We introduce an extreme difference ratio component to thoroughly mine the information in the similarity matrix and fuse it with the ranking probability, thereby effectively reweighting the original similarity matrix.

Finally, the widely utilized traditional triplet loss disregards the absolute distance between matching images and text \cite{yuan2022exploring}, \cite{yuan2022remote}, \cite{faghri2017vse++}. 
To address this quandary, we propose an optimized triplet loss that introduces an intra-class distance term, thereby effectively minimizing the distance between matching images and text.

In summary, our principal contributions are as follows:

\begin{enumerate}
\item To effectively capture the multi-scale features inherent in remote sensing imagery, we propose the Gswin transformer block, a novel architectural component. 
The proposed Gswin transformer block comprises upper and lower branches that perform cross-attention on different local window contents and the same global window, respectively, thereby effectively fusing global semantic information and local fine-grained information, facilitating the extraction of deep-level feature representations from remote sensing imagery.

\item We propose a cross-modal pre-alignment architecture predicated on global and local information from remote sensing images. 
This architecture pre-aligns images and text prior to modal fusion, thereby facilitating better cross-modal alignment and enabling more efficient subsequent fusion. 
The pre-alignment mechanism mitigates the training difficulty associated with modal fusion, consequently enhancing retrieval accuracy.

\item To fully exploit the rich information contained in the original similarity matrix, we devise a similarity matrix reweighting reranking algorithm. The algorithm introduces an extreme difference ratio component to uncover and mine significant differences in the similarity matrix. 
This component is then fused with the ranking probability to reweight the original similarity matrix, thereby effectively improving ranking quality and retrieval accuracy.

\item We propose an optimized triplet loss with an intra-class distance term. The optimized triplet loss not only minimizes the relative distance between non-matching pairs but also minimizes the absolute distance within matching pairs, thereby promoting tight clustering of same-class samples and enhancing the model's discriminative ability.
\end{enumerate}

To comprehensively evaluate the proposed CMPAGL method, extensive experiments were conducted on four publicly available remote sensing image-text datasets, namely RSICD \cite{lu2017exploring}, RSITMD \cite{yuan2022exploring}, UCM-Captions \cite{qu2016deep}, and Sydney-Captions \cite{qu2016deep}.
The experimental results substantiate that our proposed CMPAGL method exhibits superior performance compared to state-of-the-art approaches, thereby corroborating the efficacy of this architecture.

Based on the proposed methods, the subsequent content of this paper is organized as follows. 
Section \ref{Related Works} briefly reviews existing related work in the domain of RSCTIR.
Section \ref{Method} elucidates the proposed methods in detail.
Section \ref{Experiments} presents extensive comparative and ablation experiments, visualizing some results to thoroughly elucidate the effectiveness of our proposed methods. 
Lastly, Section \ref{Conclusion} summarizes the entire paper.

\section{Related Works}
\label{Related Works}
\subsection{Cross-Modal Image-Text Retrieval}  
Cross-modal image-text retrieval has long been a prevalent and challenging research topic in the multimedia domain.
Due to the significant heterogeneity and modality discrepancies between image and text data, effectively modeling and fusing the semantic information of the two modalities to achieve precise cross-modal retrieval has become a research focus and quandary. 
To this end, various methods have been proposed and extensively studied. 
You et al. \cite{you2018end} proposed an end-to-end convolutional semantic embedding model that encodes images and text into a common semantic space and trains the model by maximizing the similarity discrepancy between positive and negative sample pairs. 
Wang et al. \cite{wang2021cross} utilized deep and shallow CNNs to extract image and text features respectively, and designed specialized loss functions to enhance semantic representation. 
Additionally, some other works \cite{zheng2020dual}, \cite{zhen2019deep} followed analogous approaches. 
However, these traditional CNN/RNN-based methods grapple with effectively addressing the heterogeneity quandary between visual and language modalities. 
To overcome this quandary, some other works \cite{huang2018learning}, \cite{lee2018stacked}, \cite{xu2020cross}, \cite{li2019visual}, \cite{li2022image} adopted object detection techniques to extract features from regions of interest in images, and then fused them with text features for semantic matching. 
Although this two-stage method can achieve finer-grained object-level semantic alignment, the networks utilized for image feature extraction have low computational efficiency and are time-consuming.
With the advent of transformer models demonstrating superior modeling capabilities in natural language processing tasks, a series of transformer-based cross-modal models have emerged \cite{lu2019vilbert}, \cite{tan2019lxmert}, \cite{gao2020fashionbert}, \cite{chen2021learning}, \cite{huang2020pixel}, \cite{radford2021learning}, \cite{cheng2022vista}, \cite{kim2021vilt}, aiming to better explore the deep semantic associations between images and text. 
For instance, Huang et al. \cite{huang2020pixel} proposed directly modeling image pixels and text, eschewing intermediate feature extraction; Radford et al. \cite{radford2021learning} completely eschewed visual modality supervision and relied on text descriptions to learn visual representations; Kim et al. \cite{kim2021vilt} further simplified the visual encoder by removing convolutions and object detection; Li et al. \cite{li2021align} first aligned visual-semantic features utilizing contrastive losses before fusion; Bao et al. \cite{bao2022vlmo} designed a Mixture-of-Modality-Experts module to better encode and fuse the two modalities. These works have significantly expanded the capability of cross-modal semantic representation.

\subsection{Remote Sensing Cross-Modal Image-Text Retrieval} 
\subsubsection{CNN-based Methods} 
In remote sensing image-text retrieval tasks, CNNs have long been extensively utilized for image feature extraction, while GRUs and LSTMs are typically employed for text feature extraction. 
CNNs excel at capturing local patterns and spatial features in images, which is pivotal for understanding intricate remote sensing scenes.
Yuan et al. \cite{yuan2022exploring} proposed a multi-scale visual self-attention module and a triplet loss function to extract salient features and mitigate positive sample ambiguity; the same authors \cite{yuan2021lightweight} proposed a lightweight cross-modal retrieval model and introduced a knowledge distillation-based hidden supervision optimization method. 
The authors \cite{yuan2022remote} also proposed a multi-level information dynamic fusion (MIDF) module to effectively integrate global and local features. Ji et al. \cite{ji2023knowledge} extracted image features through CNN feature fusion and momentum contrastive learning, and performed knowledge alignment. 
Zhang et al. \cite{zhang2023hypersphere} devised an adaptive alignment strategy predicated on curriculum learning to personalize the processing of sample pairs with different difficulties, better matching images and text. 
Zhang et al. \cite{zhang2023exploring} proposed a mask-guided relation modeling and entity loss framework, utilizing a transformer encoder to capture intra-modal relations. Other related works have also adopted CNN-based methods, continuously improving retrieval performance by devising novel network modules and loss functions.

\subsubsection{Transformer-based Methods} 
Transformer-based methods have been extensively applied to cross-modal remote sensing image-text retrieval. The transformer's self-attention mechanism can directly model long-range dependencies between arbitrary positions, which is pivotal for analyzing intricate remote sensing images; it also allows seamless fusion of multiple modalities within a single architecture, rendering it well-suited for cross-modal retrieval tasks; embeddings of different modalities can also be flexibly integrated into a unified representation.
Wang et al. \cite{wang2022multi} adopted lightweight feature learning and a multi-scale interactive transformer encoder.
Yuan et al. \cite{yuan2023parameter} proposed a parameter-efficient transfer learning framework to address the need for full model fine-tuning. Chen et al. \cite{chen2024integrating} introduced a symmetric multi-level guidance network, devising local-global feature fusion guidance and fine-grained cross-modal bidirectional guidance to optimize common semantic space learning. Chen et al. \cite{chen2023multiscale} also proposed a multi-scale salient image-guided text alignment method, integrating multi-scale learning, saliency learning, and image-guided alignment mechanisms to capture fine-grained region-text correspondences. Tang et al. \cite{tang2023interacting} treated remote sensing images and text as a whole, simultaneously modeling cross-modal intrinsic relations and enhancing visual feature expression through an information interaction enhancement module. The aforementioned methods fully leverage the transformer's self-attention mechanism and cross-modal fusion capabilities, combined with various module and loss function designs, such as multi-scale fusion, interaction enhancement, contrastive learning, etc., to ameliorate the overall performance of remote sensing image-text retrieval.

In summary, whether CNN-based or Transformer-based, these methods aim to effectively extract semantic features from remote sensing images and text, and achieve more fine-grained cross-modal semantic matching to accomplish remote sensing image-text retrieval tasks, thereby making certain progress.

\section{Method}
\label{Method}
\begin{figure*}[!t]
	\centering
	\includegraphics[width=\textwidth]{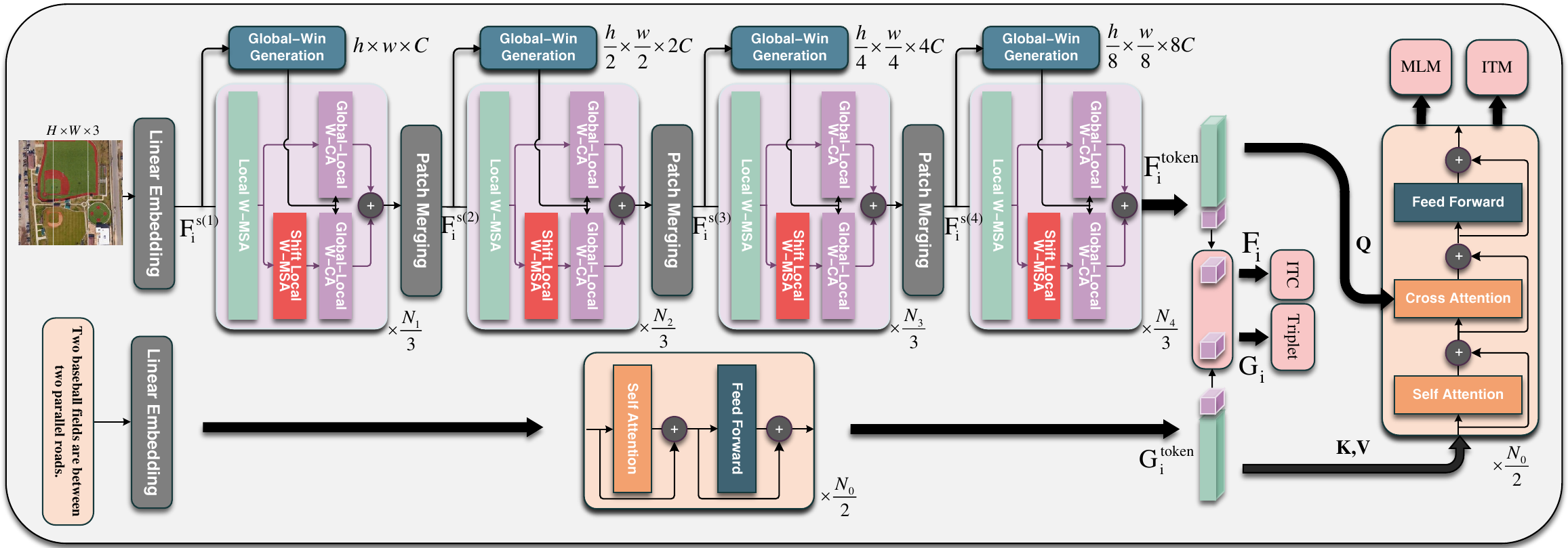}
	\caption{
		Overview of the CMPAGL architecture. The proposed framework comprises three principal components: a Gswin-based image encoder, a text encoder, and a multi-modal encoder for cross-modal feature fusion. The text encoder and multi-modal encoder share a single BERT model, with the first half ($\frac{N_0}{2}$ layers) serving as the text encoder and the latter half ($\frac{N_0}{2}$ layers) as the multi-modal encoder. After feature extraction, the model employs ITC and optimized triplet loss for pre-alignment, followed by modality fusion through the multi-modal encoder using ITM and MLM loss functions.
	}
	\label{Structure}
\end{figure*}

\subsection{Model Architecture}

As depicted in Fig.~\ref{Structure}, our proposed cross-modal pre-aligned retrieval model for global and local information (CMPAGL) facilitates the retrieval of remote sensing images and associated text by incorporating both global and local information. The model comprises three primary components: a visual encoder, a text encoder, and a multimodal encoder. Consider a dataset consisting of $N$ image-text pairs $(\mathbf{I}, \mathbf{T})$, where $\mathbf{I} = {I_1, \dots, I_i, \dots, I_N}$ represents the set of images, and $\mathbf{T} = {T_1, \dots, T_i, \dots, T_N}$ denotes the corresponding textual descriptions.
Firstly, the visual encoder maps the set of images $\mathbf{I}$ to a $d$-dimensional feature space, resulting in a set of visual features denoted as $\mathbf{F}={F_1,\dots,F_i,\dots,F_N}$.
Secondly, the text encoder maps the set of textual descriptions $\mathbf{T}$ to a $d$-dimensional feature space, denoted as $\mathbf{G}={G_1,\dots,G_i,\dots,G_N}$, representing the textual features.
Finally, notably, prior to inputting the visual features $\mathbf{F}$ and textual features $\mathbf{G}$ into the multimodal encoder, we perform a pre-alignment operation between the two modalities. This pre-alignment step ensures better alignment between the visual and textual features, thereby facilitating efficient modality fusion and further mitigating the semantic gap between the two modalities. Subsequently, the multimodal encoder takes the aligned visual features $\mathbf{F}$ and textual features $\mathbf{G}$ as input, and performs feature fusion across the two modalities.
To optimize the retrieval performance, we leverage several loss functions, encompassing image-text contrastive learning (ITC) loss and an optimized triplet loss, which are applied prior to modality fusion to pre-align the image and text features. 
Additionally, subsequent to modality fusion, we employ masked language modeling (MLM) and an image-text matching loss (ITM) to further enhance the shared semantic representations.
Furthermore, we propose a similarity matrix reweighting (SMR) reranking algorithm to refine the retrieval results. 
Next, we provide a detailed explanation of the proposed method.

\subsection{Image Feature Embedding}
This section primarily elucidates the visual encoder employed for image feature embedding. 
As depicted in Fig.~\ref{Structure}, our model performs a series of operations on the input image to extract salient visual features, which can be divided into two distinct steps. 
Initially, we perform Linear Embedding on the input image. 
Subsequently, through four stages of the proposed Gswin transformer blocks, we conduct deep feature extraction with both global and local information, thereby completing the image feature embedding process. The specific process is delineated as follows:

Firstly, we divide the input image into patches, where each patch consists of $n_{psize}$ pixels, resulting in $h \times w$ patches in total. These patches are then embedded into a C-dimensional space. Here, $h=\frac{H}{n_{psize}}$ and $w=\frac{W}{n_{psize}}$, where $H$ and $W$ denote the height and width of the input image, respectively.
In this manner, we can treat each local region of the image as an independent patch, with each patch containing a portion of the image information.
The Linear Embedding for the image  ${I_i}$ can be expressed as:
\begin{equation}
	\label{embed}	
	\mathbf{F_i^{s(1)}} = \mathbf{W_e} \cdot \mathbf{I_i}         
\end{equation}
where $\mathbf{W_e}$ denotes the linear embedding matrix of the image, and $\mathbf{F_{i}^{s(1)}} \in \mathbb{R}^{h \times w \times C}$ represents the input to the first layer of the Gswin transformer block.

Subsequently, we treat each patch as a token and input it to the Gswin transformer block.
Within this block, we perform global-local attention computation and obtain features with both global and local information, enabling our model to consider the overall structure as well as the local details of the image.
As the network layer deepens, we employ the patch merging operation between the two stages to halve the spatial dimensions of the feature map while doubling its channel depth. This operation allows our model to reduce the resolution of the feature map while preserving the feature information, thereby enhancing computational efficiency.
The specific process can be expressed as:
\begin{equation}
	\label{PM}	
	\mathbf{F_i^{s(k+1)}} = \mathbf{PM(Gswin(F_i^{s(k)}))}       
\end{equation}
where PM denotes patch merging, and k signifies the current layer of the Gswin transformer block.

After progressing through four stages of Gswin transformer blocks, we obtain the final features $\mathbf{F^{token}_i \in \mathbb{R}^{h^\prime \times w^\prime \times 8C} }$, where $h^\prime=\frac{h}{8}$ and $w^\prime=\frac{w}{8}$.
This process enables our model to extract more profound features and enhance the performance of model. 
Ultimately, we take the 0-th token as the extracted image feature $\mathbf{F_i} \in \mathbb{R}^{1 \times 1 \times d}$, which is the final output of our model. This can be expressed as:
\begin{equation}
	\begin{aligned}
		\label{Fi}	
		\mathbf{F^{token}_i} &= \mathbf{Gswin(F_i^{s(4)})}  \\
		\mathbf{F_i} &=  \mathbf{F^{token}_i}  \text{[} 0 \text{]}
	\end{aligned}
\end{equation}

In the following two subsections, we will elucidate the two core proposed modules for image feature embedding, i.e., global window generation and Gswin transformer block in detail.

\begin{figure}[htbp]
	\centering
	\subfloat[]{
		\label{fig:subfig_a}
		\includegraphics[width=\linewidth]{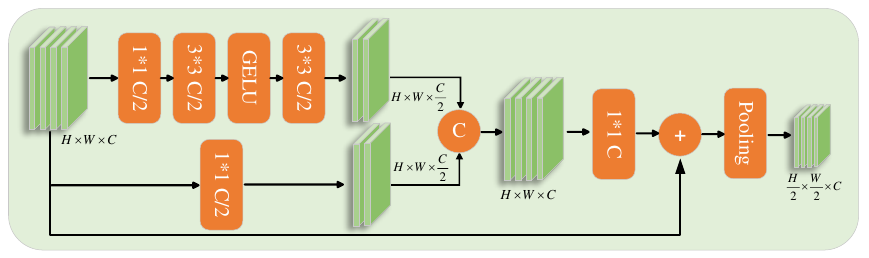}
	}
	
	\subfloat[]{
		\label{fig:subfig_b}
		\includegraphics[width=\linewidth]{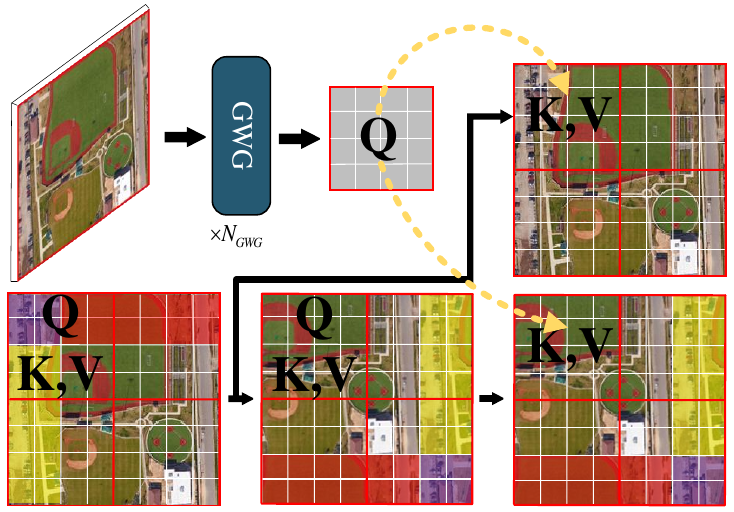}
	}
	\caption{
		(a) GWG block. (b) Schematic diagram of the interaction between the global window and the local window in the Gswin transformer block.
	}
	\label{GWG}
\end{figure}

\subsubsection{Global Window Generation}
This module primarily generates windows $\mathbf{F_w}$  with global information.
These windows are designed to have the same dimensions as the local windows, specifically $H_g = H_l$ and $W_g = W_l$, where $H_g$ and $W_g$ represent the height and width of the global windows, and $H_l$ and $W_l$ represent the height and width of the local windows, thereby facilitating the computation of cross-attention.
The specific generation network is illustrated in Fig.~\ref{fig:subfig_a}, where the image or feature map passes through GWG module of  $N_{GWG}$ layers, extracting the global features of the remote sensing image, and ultimately generating windows with global information.

When taking $\mathbf{F_i^{s(k)}}$ as input, the process of inputting to GWG can be expressed as:
\begin{equation}
	\label{e1}
	\begin{aligned}
		\mathbf{x_1} &= \operatorname{Conv}_{3 \times 3}^{(2)}\left(\operatorname{GELU}\left(\operatorname{Conv}_{3 \times 3}^{(1)}\left(\operatorname{Conv}_{1 \times 1}^{(1)}(\mathbf{F_i^{s(k)}})\right)\right)\right) \\
		\hat{\mathbf{x}} &= \operatorname{Concat}\left(\mathbf{x_1},\operatorname{Conv}_{1 \times 1}^{(2)}(\mathbf{F_i^{s(k)}})\right)
	\end{aligned}
\end{equation}
where $\operatorname{Conv}{1 \times 1}^{(i)}$ and $\operatorname{Conv}{3 \times 3}^{(i)}$ represent the $i$-th $1 \times 1$ and $3 \times 3$ convolution operations respectively, ${\mathbf{x_1}}$ is the intermediate feature map, $\mathbf{F_i^{s(k)}}$ is the input feature map, and GELU represents the Gaussian Error Linear Unit activation function. 
The first $1 \times 1$ convolution operation ($\operatorname{Conv}{1 \times 1}^{(1)}$) is applied to extract initial features from the input. Subsequently, two $3 \times 3$ convolution operations ($\operatorname{Conv}{3 \times 3}^{(1)}$ and $\operatorname{Conv}_{3 \times 3}^{(2)}$) are employed to expand the receptive field of the network. These multi-layer nonlinear convolution layers can further extract more intricate features.
Incorporating the GELU activation function between the two $3 \times 3$ convolution operations enhances the nonlinearity of the network and accelerates the convergence of the model.
The second $1 \times 1$ convolution operation ($\operatorname{Conv}_{1 \times 1}^{(2)}$) is applied directly to the input $\mathbf{F_i^{s(k)}}$ to extract additional feature information while preserving the original features.
Finally, the outputs from both paths are concatenated along the channel dimension to yield $\hat{\mathbf{x}}$, combining the processed features with the preserved original information.

The above Eq.\ref{e1} integrates features at different levels, thereby enabling the extraction of richer and more representative features. 
Additionally, downsampling is required to generate a global window of the same size as the local window. This process can be expressed as:
\begin{equation}
\label{e2}	
\mathbf{F_w} = \operatorname{Pooling}(\operatorname{Conv}_{1 \times 1}^{(3)}(\hat{\mathbf{x}}) + \mathbf{F_i^{s(k)}})
\end{equation}
where $\mathbf{F_w}$ is the generated global window feature map, $\operatorname{Conv}_{1 \times 1}^{(3)}$ represents the third $1 \times 1$ convolution operation.
The pooling layer reduces the input length and width to half of their original sizes.
In this manner, Eq.\ref{e1} and Eq.\ref{e2} constitute a complete GWG block.
By stacking $N_{GWG}$ layers, the input containing global information is generated into a global information window with the same size as the local window. 
The value of $N_{GWG}$ is determined by $H_l $ and $W_l$ of the local window and the height and width of the input feature $\mathbf{F_i^{s(k)}}$.
\subsubsection{Gswin Transformer Block}
The Gswin transformer block constitutes the core module of our model, which can effectively leverage global and local information to enhance the performance of cross-modal retrieval for remote sensing images and associated text.
Specifically, it includes the following steps:

First, Local W-MSA, a local window multi-head self-attention mechanism, is performed on the input data by calculating the attention weights within each local window, yielding the local feature $\mathbf{F_{L}}$, which captures detailed information of the local regions. Subsequently, $\mathbf{F_{L}}$ is divided into two branches, where different global and local interaction operations are performed. The first branch applies Global-Local W-CA, a global-local window cross-attention mechanism, on $\mathbf{F_{L}}$ and $\mathbf{F_{W}}$, where $\mathbf{F_{W}}$ is generated by the GWG module. This operation is conducted on the original (non-shifted) local windows, aligning and integrating global features with the corresponding local information. The second branch first performs Shift Local W-MSA, a shifted local window multi-head self-attention mechanism, calculating attention weights within each shifted local window to obtain the shifted interaction information $\mathbf{F_{SL}}$, which enhances the relational information between neighboring local regions. Subsequently, Global-Local W-CA is performed on $\mathbf{F_{SL}}$ and the same global window $\mathbf{F_W}$, integrating global features with the updated local information from the shifted windows. 
Finally, the outputs of the two branches, $\mathbf{F_{GL1}}$ and $\mathbf{F_{GL2}}$, are summed to yield the final local feature representation $\mathbf{F_{GL}}$. By combining the outputs from both original and shifted local windows, the model effectively captures complementary details from adjacent regions, enriching the feature representation and providing a more comprehensive understanding of the spatial context. 

\begin{figure}[!t]
	\centering
	\includegraphics[width=\linewidth]{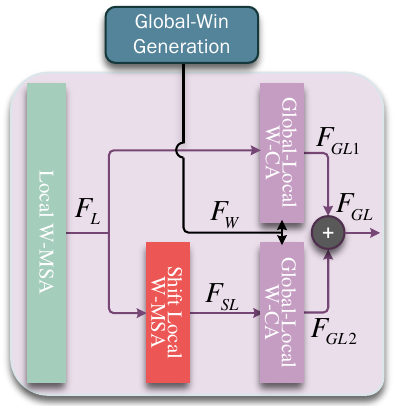}
	\caption{Gswin transformer block. This method fully combines global and local features through global-local window attention.}
	\label{GswinBlock}
\end{figure}

Fig.~\ref{fig:subfig_b} illustrates the attention interaction process between the global window and the \textit{shift} local windows in the Gswin transformer block. The structure of the Gswin transformer block is illustrated in Figure \ref{GswinBlock}, which can be mathematically expressed as:
\begin{equation}
	\label{e3}	
	\begin{aligned}
		\mathbf{F_L} &= \operatorname{LW-MSA}(\mathbf{F_i^{s(k)}}) \\
		\mathbf{F_{SL}} &= \operatorname{SLW-MSA}(\mathbf{F_L}) \\
		\mathbf{F_{GL}} &= \operatorname{GLW-CA}(\mathbf{F_L},\mathbf{F_W}) + \operatorname{GLW-CA}(\mathbf{F_{SL}},\mathbf{F_W})
	\end{aligned}
\end{equation}

This approach offers multiple advantages:
On the one hand, local windows may separate objects of the same category, leading to incomplete semantic information of the objects. 
By employing the shifted window form, the semantic information of the objects can be supplemented, thereby improving the quality of the local features.
On the other hand, by performing attention with different contents of local windows and the same global window, multi-scale features can be better captured, thereby enhancing the generalization ability of the model and enabling better local image-text matching for small targets in the image described by the text.

\subsection{Text Feature Embedding}
This section primarily introduces the text encoder employed for text feature embedding.
To accurately capture text features, the Bert model with a Masked Language Model (MLM) \cite{devlin2018bert} is adopted as the text encoder. 
Within the CPAGL architecture,  MLM can fully leverage image information for word cloze, thereby enhancing the similarity between the matched text and image features, further bridging the semantic gap between modalities.
Inspired by VILT \cite{liu2021swin} and ALBEF \cite{li2021align}, we recognize that an excellent multi-modal model necessitates a large visual encoder, whereas the text encoder need not be excessively intricate.
Consequently, we partition the $N_0$-layer Bert model, using the first $\frac{N_0}{2}$ layers as the text encoder and the last $\frac{N_0}{2}$ layers as a multi-modal encoder. This design not only saves parameters but also achieves good performance.
The multi-modal encoder will be introduced in the subsequent Section \ref{Modality Interaction}. 
Let the query text $T_i$ be composed of $N_T$ words, which can be represented as: $T_i={w_{1}^i, \dots ,w_{t}^i,\dots,w_{N_T}^i}$, where $w_{t}^i$ denotes the $t$-th word of the $i$-th text.
The words are then embedded into word vectors, which can be represented as:
\begin{equation}
	\label{e4}	
	\mathbf{V_i} =  \operatorname{Embed}(T_i)
\end{equation}
where $\operatorname{Embed}(\cdot)$ denotes the word embedding function, and $\mathbf{V_i}$ represents the feature matrix after text encoding.
 
Subsequently, the feature matrix $V_i$ is input into the $\frac{N_0}{2}$ layer text encoder to obtain the feature $\mathbf{G^{token}_i}$.
\begin{equation}
	\label{e5}	
	\mathbf{G^{token}_i} =  \operatorname{Bert}_{text}(V_i)
\end{equation}

Finally, the 0-th token of $\mathbf{G^{token}_i}$ is selected as the final text feature $\mathbf{G_i} \in \mathbb{R}^{1 \times 1 \times d}$, a process akin to the selection of image features:
\begin{equation}
	\label{G0}	
	\mathbf{G_i}=\mathbf{G^{token}_i}\text{[} 0 \text{]}
\end{equation}

\subsection{Modality Interaction}
\label{Modality Interaction}
This section concentrates on multimodal encoders for modal interaction.
Modal interaction refers to the exchange of information between diverse modalities (such as images and text). 
This is crucial in cross-modal retrieval as it enables the model to comprehend and leverage the associations between images and text. 
Through effective modal interaction, models can better comprehend and represent cross-modal data, consequently enhancing retrieval performance.
However, if the features are not well aligned prior to being input to the multi-modal encoder, the fusion between cross-modal features can become exceedingly arduous. 

To address this issue, we adopt the pre-align strategy. 
As illustrated in Fig.~\ref{Structure}, we align the image features $\mathbf{F_i}$, extracted by the image encoder, with the text features $\mathbf{G_i}$, extracted by the text encoder, before they enter the multimodal encoder. This alignment is achieved using image-text contrastive (ITC) loss and optimized triplet loss, ensuring a level of consistency between the image and text features that enhances the effectiveness of the subsequent modal fusion process.

Upon completing the pre-alignment, we subsequently input $\mathbf{F^{token}_i}$ and $\mathbf{G^{token}_i}$ conjointly into the multi-modal encoder for modality cross attention, fully integrating the modal information.
The final output subsequent to fusion is utilized for ITM and MLM calculation. 
ITC, ITM, MLM and optimized triplet losses will be elucidated in the subsequent Section \ref{Objective Function}.

\subsection{Similarity Matrix Reweighting (SMR) Rerank}

Upon feature extraction utilizing the methodologies delineated in the preceding sections, we can extract features of $\mathbf{N_i}$ images and $\mathbf{N_t}$ texts, and obtain an original similarity matrix $S_{raw} \in \mathbb{R}^{N_i \times N_t}$.
Subsequently, $S_{raw}$ can be sorted to procure the top $\mathbf{K}$ retrieval candidate images of text $T_i$ (or the top $\mathbf{K}$ retrieval candidate texts of image $I_i$). 
However, the original similarity matrix $S_{raw}$ at this juncture does not fully consider the intrinsic connection between two-way retrieval, owing to the fact that after the image and text are successfully matched, they can retrieve each other. 
Grounded on the aforementioned, Wang et al. \cite{wang2019matching} proposed a cross-modal re-ranking algorithm to perform reverse retrieval of the top $\mathbf{K}$ candidates, and inspect the position ordering in the reverse retrieval to procure the final retrieval results.
Yuan et al. \cite{yuan2022remote} grounded on Wang's work, incorporated the significance component to calculate the proportion of reverse retrieval candidate similarity to the total. 
Although Yuan further utilizes the original similarity matrix information in comparison to Wang, it does not consider the probability information of $S_{raw}$. 
Instead, it directly calculates the score in accordance with the retrieval ranking to procure the final reranked similarity matrix. 

\begin{figure}[!t]
	\centering
	\includegraphics[width=\linewidth]{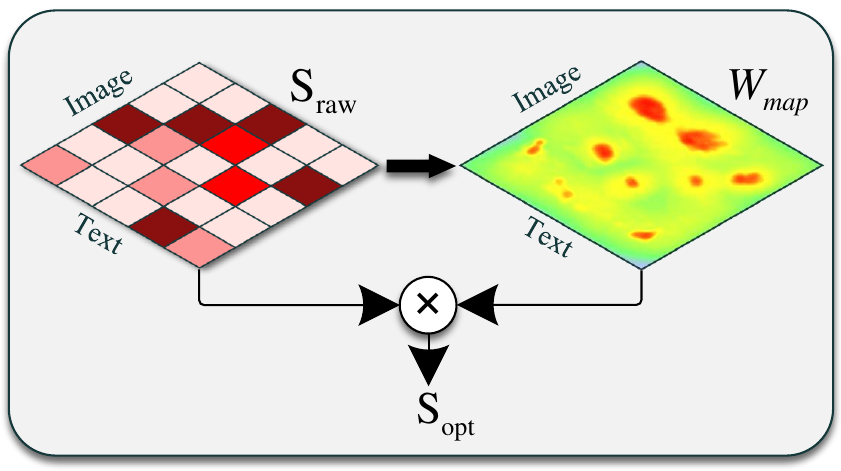}
	\caption{
		Schematic diagram of the re-ranking algorithm. In this schematic, we present a re-ranking algorithm for image and text retrieval. Firstly, we start from the original image and text similarity matrix $S_{raw}$. Based on the similarity information, we compute the re-ranking probability weighted matrix $W_{map}$. Next, we reweight $S_{raw}$ to obtain the optimized similarity matrix $S_{opt}$.
	}
	\label{rerank}
\end{figure}

In order to fully leverage the original $S_{raw}$ similarity matrix information, we propose a novel reranking algorithm called similarity matrix reweighting (SMR) rank, as illustrated in Fig.~\ref{rerank}.
We obtain the ranking information from the original similarity matrix, and convert it into probability through the ranking information from i2t or t2i forward or reverse retrieval, with higher ranks corresponding to higher probabilities. 
Additionally, we proposed the extreme difference ratio component to fully exploit the information in the original similarity matrix $S_{raw}$. 
In contrast to Yuan's significance component, which calculates the ratio of a single similarity score to the sum of similarities in the entire column (or row) of the similarity matrix, our extreme difference ratio component calculates the ratio of a single similarity score to the maximum similarity score in both forward and reverse retrieval. This ratio of maximum similarities can better capture and exploit salient information.
Subsequently, the probability of ranking information conversion is amalgamated with the extreme difference ratio component to obtain the similarity probability weighted matrix $W_{map}$.
The original similarity matrix $S_{raw}$ is reweighted to procure the final optimized similarity matrix $S_{opt}$. 

To provide a detailed explanation of our cross-modal retrieval reranking algorithm, we will illustrate it with an example from the perspective of image-to-text retrieval.

Firstly, we utilize image $I_q$, the query image, to retrieve the top $\mathbf{K}$ similar texts, defined as:

\begin{equation}
	\label{e6}	
	R_{i2t}(i,K) = \{t_1,\dots,t_j,\dots,t_K\}
\end{equation}
where $t_j$ represents the j-th text that is most similar to the query image. 
We utilize forward retrieval ranking information and define the forward ranking component as:
\begin{equation}
	\label{e7}	
	w_{i2t} = 1-\frac{j}{K}
\end{equation}

The objective of the aforementioned formula is to convert the search ranking into probability.
The higher the ranking, the greater the probability. 
We utilize linear weights to transform probabilities, which are better able to preserve larger differences between rankings than exponential weights, thus better capturing the relative significance of rankings. 
Owing to the exponential function decaying at a faster rate, the weight of the lower rankings is excessively low. 
Secondly, due to the inherent characteristics of the exponential function, the weight difference between the top rankings is relatively small. 
In contrast, linear weighting assigns weights based on the precise gaps in rankings.

Subsequently, we need to utilize one of the top $\mathbf{K}$ texts obtained above to perform reverse retrieval, and then inspect the position of the query image $I_q$ and convert it into a reverse ranking component.
The reverse retrieval step is defined as:
\begin{equation}
	\label{e8}	
	R_{t2i}(t_j,N) = \{i_1,\dots,i_k,\dots,i_N\}
\end{equation}
where $i_k$ denotes the k-th image most similar to the query text $t_j$in reverse retrieval. 
$\mathbf{N_i}$ denotes the total number of images.
At this juncture, we find the position of the query image $I_q$ in the reverse retrieval. The reverse ranking component is defined as follows:
\begin{equation}
	\label{e9}	
	w_{t2i} =  1-\frac{k}{N}
\end{equation}
The amalgamation of the forward retrieval component $w_{i2t}$ and reverse retrieval component $w_{t2i}$ can comprehensively leverage text queries and image features to perform more comprehensive cross-modal retrieval and enhance the relevance and diversity of retrieval results.

In order to further exploit the information in the original similarity matrix, we proposed the extreme difference ratio component to further unearth the deep-level salient components. 
It is defined as follows:
\begin{equation}
	\label{e10}	
	w_{md} =  \frac{s(t_j,i_k)}{\text{Max}_{\text{p1}}^{row} (S_{raw})} + \frac{s(t_j,i_k)}{\text{Max}_{\text{p2}}^{col}(S_{raw})}
\end{equation}
where $s(t_j,i_k)$ denotes the similarity between $t_j$ and $i_k$, $\text{Max}_{\text{p1}}^{row}$ is utilized to retrieve the maximum value from row $p_1$ of $S_{raw}$, with $p_1=pos(i_k)$ indicating the original position of $i_k$ in $S_{raw}$. 
On the other hand, $\text{Max}_{\text{p2}}^{col}$ is utilized to retrieve the maximum value from column $p_2$ of $S_{raw}$, with $p_2=pos(t_j)$ indicating the position of $t_j$ in $S_{raw}$. 
The extremal difference ratio component primarily reflects the discrepancy between the similarity $s(i_q,t_j)$ of the current query image $i_q$ and the retrieval text $t_j$, and the similarity $\text{Max}_{p_1}^{row}$ between $i_q$ and the most matching text. 
In other words, it accentuates the difference between $s(i_q,t_j)$ and $\text{Max}_{p_1}^{row}$. 
Similarly, it also captures the discrepancy between $s(i_q,t_j)$ and the similarity $\text{Max}_{p_2}^{col}$ of the best matching image acquired during reverse retrieval of text $t_j$. 
That is, it accentuates the difference between $s(i_q,t_j)$ and $\text{Max}_{p_2}^{col}$.

Ultimately, our similarity probability weighted matrix $W_{map}$ can be expressed as:
\begin{equation}
	\label{e11}	
	W_{map} =  w_{i2t} + \gamma_1 \cdot w_{t2i} + \gamma_2 \cdot w_{md}
\end{equation}
where $\gamma_1$ and $\gamma_2$ are weighting coefficients, the final weighted optimized similarity matrix $S_{opt}$ can be expressed as:

\begin{equation}
	\label{e12}	
	S_{opt} =   W_{map} \cdot S_{raw}
\end{equation}

\subsection{Objective Function}
\label{Objective Function}
The preceding sections primarily delineated the specific architecture of the proposed model. In this section, we shall elucidate in detail the objective functions employed in the pre-alignment stage and modal interaction.
To facilitate effective retrieval between remote sensing images and text, we employ image-text contrastive learning loss and optimized triplet loss for pre-alignment prior to the multi-modal encoder, thereby simplifying modal fusion.
The image-text contrastive loss primarily focuses on the global distribution information of images and texts, potentially overlooking local semantic information and detail matching.
Hence, we propose an optimized triplet loss to complement the local or detailed contrastive features between images and texts.
Subsequent to the multi-modal encoder, we employ masked language modeling (MLM) and image-text matching (ITM) for joint training.
In contrast to BERT's single-modal training, our MLM can fully exploit image information to perform cloze tasks, enabling the model to concurrently learn the semantic information and relationship information of images and texts, thereby achieving more fine-grained image and text alignment.
The overall loss function of our model, denoted as $\mathcal{L}$, is formulated as:
\begin{equation}
	\begin{aligned}
		\mathcal{L} = \mathcal{L}_{ITC}+\mathcal{L}_{Triplet}+\mathcal{L}_{MLM}+\mathcal{L}_{ITM}
	\end{aligned}
\end{equation}
In the following subsections, we will provide a detailed explanation of each loss function and its specific contribution to the overall model.

\subsubsection{Image-Text Contrastive Learning }
To attain effective alignment between images and text, we employ image-text contrastive learning (ITC) as our first objective function.
ITC is a loss function predicated on momentum contrastive learning, which can use the momentum model \cite{he2020momentum} to maintain a global feature repository of images and texts, thereby enhancing the global distribution information between images and texts. 
The core tenet of momentum contrastive learning is to employ a slow-updating momentum encoder to generate negative samples, thereby circumventing the computational overhead of reconstructing the feature repository at each iteration. 
We apply the tenet of momentum contrastive learning to cross-modal scenarios of images and texts, thereby enabling our model to learn meaningful feature representations on a large corpus of image-text pair data.
It is defined as follows:
\begin{figure}[!t]
	\centering
	\includegraphics[width=\linewidth]{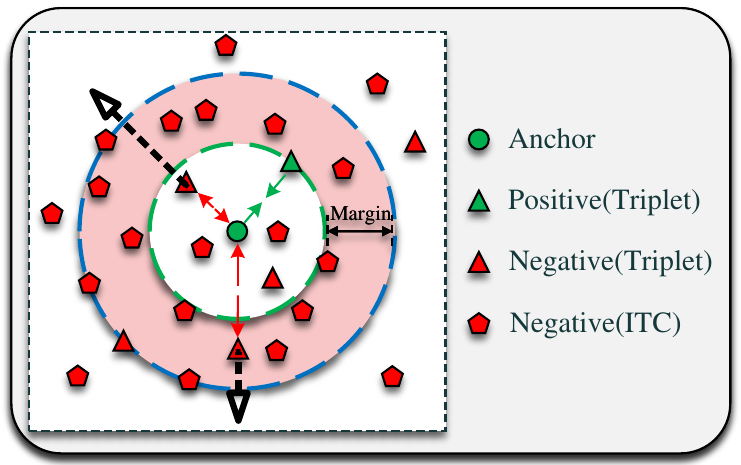}
	\caption{
		Optimized triplet loss positive and negative sample optimization diagram. The goal of this loss function is not only to ensure that the distance between matching image-text pairs and unmatched image-text pairs is at least greater than the preset Margin, but also to pursue the ultimate narrowing of the distance between matching image-text pairs. The figure also shows that optimized triplet loss focuses on local samples, while ITC loss can focus on more samples because of the dynamically updated momentum encoder, allowing comparison from a global perspective.}
	\label{Loss}
\end{figure}
\begin{equation}
	\begin{aligned}
		&\mathcal{L}_{ITC} = -\frac{1}{2N}\sum_{i=1}^N \Biggl[\log \frac{\exp(\text{S}(I_i, T_i)/\tau)}{\sum_{j=1}^{M}\exp(\text{S}(I_i, \tilde{T}_j)/\tau)} \Biggr.\\
		&+ \Biggl. \log \frac{\exp(\text{S}(T_i,I_i)/\tau)}{\sum_{j=1}^{M}\exp(\text{S}({T}_i, \tilde{I}_j)/\tau)}\Biggr]
	\end{aligned}
\end{equation}
where $S(I, T)$ denotes the similarity between the image and the text, cosine similarity being employed as the similarity function. 
$N$ represents the number of image-text pairs in a batch, $M$ denotes the size of the momentum encoder, and $\tau$ signifies the temperature parameter. $\tilde{T}$ and $\tilde{I}$ represent the image and text in the momentum encoder, respectively.

\subsubsection{Optimized Triplet Loss}
The aforementioned ITC introduces momentum updates, thereby enables the model to learn more stable and consistent feature representations. 
However, ITC solely focuses on global comparisons between images and texts, potentially overlooking local semantic information or detail matching. Hence, we propose an optimized triplet loss. 
As illustrated in Fig.~\ref{Loss}, owing to triplet loss being typically computed within a batch, it can complement local or detailed contrastive features between images and text, thereby augmenting the semantic matching between them.
Traditional triplet loss solely focuses on rendering matching images and texts more similar than non-matching ones in terms of relative distance, overlooking the within-class distance between matching images and texts. 
Hence, we enhance the triplet loss to render the similarity between matching images and texts as proximate as possible to 1. 
This can ameliorate the performance of cross-modal retrieval, as higher similarity between matching images and texts leads to higher retrieval accuracy. 
The optimized triplet loss is defined as follows:
\begin{equation}
	\begin{aligned}
		&\mathcal{L}_{Triplet}=\sum_{i=1}^N \sum_{j=1}^N [\alpha -\text{S}(I_i, T_i)+\text{S} (I_i,T_j)]_{+} \\
		& +\sum_{j=1}^N \sum_{i=1}^N [\alpha-\text{S} (I_j, T_j)+\text{S} (I_i, T_j)]_{+}
		+\sum_{i=1}^N [ 1 - \text{S}(I_i,T_i)]
	\end{aligned}
\end{equation}
where $\alpha$ is the boundary parameter, $[x]_+$ denotes $\max(0, x)$.

\subsubsection{Masked Language Modeling}
The output end of the multi-modal encoder employs masked language modeling, which can fully exploit the information of the image to predict the obscured portions of the text, thereby augmenting the alignment between the image and the text. 
The core tenet of MLM is to randomly substitute some words in the text with special mask symbols, and subsequently enable the model to predict the masked words predicated on the context and image information.
It is defined as follows:
\begin{equation}
	\mathcal{L}_{MLM}=-\frac{1}{N} \sum_{i=1}^N \sum_{t=1}^{N_T} \Gamma\left(m_{t}^i=1\right) \log P\left(w_{t}^i \mid w_{1}^i, \ldots, w_{N_T}^i, I_i\right)
\end{equation}
where $N_T$ denotes the maximum length of the text, $w_{t}^i$ is the $t$-th word of the $i$-th sample, and $m_{i,t}$ is a masking mask. If the $t$-th word of the $i$-th sample is masked, then $m_{t}^i=1$, otherwise $m_{t}^i=0$. $\Gamma(x)$ is an indicator function, if $x$ is true, then $\Gamma(x)=1$, otherwise $\Gamma(x)=0$. 
$P(w_{t}^i \mid w_{1}^i,\ldots,w_{N_T}^i,I_i)$ is a conditional probability, denoting the probability of predicting the $t$-th word of the $i$-th sample given the image and other portions of the text.

\subsubsection{Image-Text Matching}
ITM is a cross-entropy-based loss function that directly optimizes the matching score between images and text, thereby ameliorating the alignment between images and text. Specifically, our ITM loss function is defined as follows:
\begin{equation}
	\begin{aligned}
	 \mathcal{L}_{ITM} = -\frac{1}{N}\sum_{i=1}^{N} [y_i \log(p_{itm}(\mathbf{F^{token}_i},\mathbf{G^{token}_i}))\\
	 + (1-y_i) \log(1-p_{itm}(\mathbf{F^{token}_i},\mathbf{G^{token}_i}))]
	 \end{aligned}
\end{equation}
where $y_i$ is the label of the $i$-th sample. If the image and text are matched, then $y_i=1$, otherwise $y_i=0$. $p_{itm}(\mathbf{F^{token}_i}, \mathbf{G^{token}_i})$ denotes the matching probability of the image-text pair given the feature vectors $\mathbf{F^{token}_i}$ and $\mathbf{G^{token}_i}$, which is computed by our multimodal encoder.

\section{Experiments}
\label{Experiments}
\subsection{Dataset and Evaluation Metrics}
To comprehensively evaluate the effectiveness of our proposed CMPAGL model on the bidirectional retrieval task of remote sensing images and text, we conducted experiments on four publicly available benchmark datasets, including RSICD, RSITMD, UCM-Captions, and Sydney-Captions. These datasets were collected from various remote sensing platforms and sources, covering diverse remote sensing scenarios and targets with high diversity and complexity. The information of the four datasets is shown in Table \ref{tab:dataset_stats}. We provide a brief introduction to these datasets as follows:

• The RSICD dataset \cite{lu2017exploring} is used for remote sensing image description generation, containing 10,921 remote sensing images with different spatial resolutions. Each image has a resolution of 224×224 pixels and is associated with five descriptive text sentences, covering various remote sensing scenarios such as farmlands, lakes, airports, bridges, cities, etc.

• The RSITMD dataset \cite{yuan2022exploring} is used for cross-modal retrieval of remote sensing images and text, including 4,743 remote sensing images spanning 32 remote sensing scenes. Each image has a resolution of 256×256 pixels and is associated with five descriptive text sentences, which contain information about the relationships and attributes of objects, such as color, shape, and location, suitable for fine-grained remote sensing image retrieval tasks.

• The UCM-Captions dataset \cite{qu2016deep} is based on the UC Merced Land Use dataset, comprising 2,100 RGB aerial images from 21 categories. Each image has a resolution of 256×256 pixels with a spatial resolution of 1 foot, and is associated with five descriptive text captions describing the main land features and characteristics in the image.

• The Sydney-Captions dataset \cite{qu2016deep} is constructed based on the Sydney dataset, including 613 remote sensing images from seven categories in Sydney, Australia. Each image has a resolution of 500×500 pixels with spatial resolutions ranging from 0.5 meters to 30 meters, and is associated with five captions summarizing the scenes and objects in the image.

We adopted the widely used evaluation metric: Recall@K (R@K), which represents the percentage of queries for which the correct result is retrieved within the top K retrieved results. We report the R@K results for three different K values (1, 5, 10). 
For remote sensing image-to-text retrieval, we treat a remote sensing image as the query and retrieve the most relevant text annotations from all available text annotations. For text-to-remote sensing image retrieval, we use a text description as the query and retrieve the most relevant remote sensing images from all available images. 
We also evaluate our model using the mean Recall (mR) metric, which is particularly useful for cross-modal retrieval tasks. mR is calculated as the average of all Recall@K (R@1, R@5, R@10) metrics for both image retrieval and text retrieval. This metric provides a comprehensive assessment of the retrieval performance across different K values and both modalities.
We compared our model with several existing methods, and the experimental results on the four datasets demonstrate that our model outperforms other methods in both directions of remote sensing image and text retrieval, proving its effectiveness and robustness.

\begin{table}[htbp]
	\centering
	\caption{Dataset Statistics}
	\label{tab:dataset_stats}
	\begin{tabular}{lcccc}
		\toprule
		Dataset & Images & Texts & Text average Length \\
		\midrule
		RSICD & 10,921 & 54,605  & 11.57  \\
		RSITMD & 4,743 & 23,715  & 11.32  \\
		UCM-Captions & 2,100 &  10,500 & 11.52 \\
		Sydney-Captions & 613 & 3,065 & 13.18 \\
		\bottomrule
	\end{tabular}
\end{table}

\begin{table*}[htbp]
	\centering
	\caption{Comparative Performance of Image-Text Retrieval on Datasets: RSICD, RSITMD, UCM, and SYDNEY}
	\label{table2}
	\resizebox{\textwidth}{!}{%
		\begin{tabular}{c|c|c|c|c|c|c|c|c|c|c|c|c|c|c}
			\specialrule{.1em}{.1em}{.1em}
			& \multicolumn{6}{c}{RSICD Dateset} & \phantom{abc} & \multicolumn{6}{c}{RSITMD Dataset} \\
			\cmidrule{1-8} \cmidrule{9-15}
			\multirow{2}{*}{Approach}& \multicolumn{3}{c}{Text Retrieval}\vline & \multicolumn{3}{c}{Image Retrieval} \vline&& \multicolumn{3}{c}{Text Retrieval}\vline & \multicolumn{3}{c}{Image Retrieval}\vline \\
			 & R@1 & R@5 & R@10 & R@1 & R@5 & R@10 & mR & R@1 & R@5 & R@10 & R@1 & R@5 & R@10 & mR\\
			\midrule
			AMFMN-soft & 5.05 & 14.53 & 21.57 & 5.05 & 19.74 & 31.04 & 16.02 & 11.06 & 25.88 & 39.82 & 9.82 & 33.84 & 51.90 & 28.74\\
			AMFMN-fusion & 5.39 & 15.08 & 23.40 & 4.90 & 18.28 & 31.44 & 16.42 & 11.06 & 29.20 & 38.72 & 9.96 & 34.04 & 52.96 & 29.32\\
			AMFMN-sim & 5.21 & 14.72 & 21.57 & 4.08 & 17.00 & 30.60 & 15.53 & 10.63 & 24.78 & 41.81 & 11.51 & 34.69 & 54.87 & 29.72\\
			LW-MCR-b & 4.57 & 13.71 & 20.11 & 4.02 & 16.47 & 28.23 & 14.52 & 9.07 & 22.79 & 38.05 & 6.11 & 27.74 & 49.56 & 25.55\\
			LW-MCR-d & 3.29 & 12.52 & 19.93 & 4.66 & 17.51 & 30.02 & 14.66 & 10.18 & 28.98 & 39.82 & 7.79 & 30.18 & 49.78 & 27.79\\
			GaLR & 6.59 & 19.85 & 31.04 & 4.69 & 18.48 & 32.13 & 18.96 & 14.82 & 31.64 & 42.48 & 11.15 & 36.68 & 51.68 & 31.41\\
			CLIP & 8.02 & 22.76 & 35.07 & 5.62 & 21.13 & 35.31 & 21.32 & 14.77 & 34.85 & 46.38 & 10.23 & 34.02 & 47.79 & 31.34\\
			IEFT & 8.43 & 27.15 & 40.74 & 7.65 & 27.35 & 42.19 & 25.86 & 15.34 & 36.67 & 50.34 & 11.02 & 37.21 & 57.31 & 34.65\\
			HVSA & 7.47 & 20.62 & 32.11 & 5.51 & 21.13 & 34.13 & 20.16 & 13.20 & 32.08 & 45.58 & 11.43 & 39.20 & 57.45 & 33.16\\
			KAMCL & 12.08 & 27.26 & 38.70 & 8.65 & 27.43 & 42.51 & 26.10 & 16.51 & 36.28 & 49.12 & 13.50 & 42.15 & 59.32 & 36.14\\
			CMPAGL w/o SMR & 11.71 & 27.08 & 41.17 & 8.86 & 29.03 & 44.72 & 27.10 & 18.81 & 36.72 & 51.11 & 14.48 & 41.65 & 60.35 & 37.18\\
			CMPAGL with SMR & \textbf{12.71} & \textbf{29.55} & \textbf{42.54} & \textbf{9.08} & \textbf{31.11} & \textbf{46.81} & \textbf{28.63} & \textbf{19.13} & \textbf{42.36} & \textbf{54.74} & \textbf{15.67} & \textbf{44.32} & \textbf{60.51} & \textbf{39.46}\\
			\midrule
			& \multicolumn{6}{c}{UCM-Captions Dataset} && \multicolumn{6}{c}{Sydney-Captions Dataset}\\
			\cmidrule{1-8} \cmidrule{9-15}
			\multirow{2}{*}{Approach}& \multicolumn{3}{c}{Text Retrieval} \vline& \multicolumn{3}{c}{Image Retrieval} \vline&& \multicolumn{3}{c}{Text Retrieval}\vline & \multicolumn{3}{c}{Image Retrieval}\vline\\
			 & R@1 & R@5 & R@10 & R@1 & R@5 & R@10 & mR & R@1 & R@5 & R@10 & R@1 & R@5 & R@10 & mR\\
			\midrule
			AMFMN-soft & 12.86 & \textbf{51.90} & 66.67 & \textbf{14.19} & 51.71 & 78.48 & 45.97 & 20.69 & 51.72 & 74.14 & 15.17 & 58.62 & 80.00 & 50.06\\
			AMFMN-fusion & \textbf{16.67} & 45.71 & 68.57 & 12.86 & 53.24 & 79.43 & 46.08 & 24.14 & 51.72 & 75.86 & 14.83 & 56.55 & 77.89 & 50.17\\
			AMFMN-sim & 14.76 & 49.52 & 68.10 & 13.43 & 51.81 & 76.48 & 45.68 & \textbf{29.31} & 58.62 & 67.24 & 13.35 & 60.00 & \textbf{81.72} & 51.72\\
			LW-MCR-b & 12.38 & 43.81 & 59.52 & 12.00 & 46.38 & 72.48 & 41.10 & 17.24 & 48.28 & 72.41 & 14.13 & 56.90 & 77.24 & 47.70\\
			LW-MCR-d & 15.24 & \textbf{51.90} & 62.86 & 11.90 & 50.95 & 75.24 & 44.68 & 18.97 & \textbf{58.63} & 75.86 & 13.45 & 57.59 & 78.97 & 50.57\\
			GaLR & 12.87 & 43.65 & 60.55 & 11.69 & 47.84 & 73.55 & 41.69 & 16.44 & 52.43 & 71.32& 14.79 & 57.16 & 77.65 & 48.30\\
			CLIP & 13.57 & 46.35 & 62.43 & 10.08 & 46.34 & 73.66 & 42.07 & 24.35 & 54.37 & 73.21 & 17.57 & 55.72& 74.39 & 49.94\\
			IEFT & 13.26 & 49.43 & 67.89 & 13.43 & 52.56 & 82.79 & 46.56 & 26.78 & 56.25 & \textbf{76.57} & 14.89 & 57.19 & 80.00 & 51.94\\
			HVSA & 12.85 & 46.03 & 68.25 & 11.90 & 47.14 & 72.48 & 43.11 & 12.64 & 40.63 & 66.32 & 13.79 & 48.39 & 70.00 & 41.96\\
			KAMCL& 13.17 & 47.25 & 68.18 & 12.35 & 49.25 & 76.33 & 44.42 & 14.32 & 44.88 & 68.57 & 15.27 & 55.18 & 73.86 & 45.34\\
			CMPAGL w/o SMR & 11.90 & 46.67 & 71.43 & 12.86 & 56.95 & 92.47 & 48.71 & 25.86 & 47.82 & 68.51 & 20.34 & 60.34 & 79.31 & 50.36\\
			CMPAGL with SMR & 12.86 & 47.62 & \textbf{72.86} & 13.52 & \textbf{58.95} & \textbf{93.05} & \textbf{49.81} & 27.32 & 49.34 & 73.32 & \textbf{21.03} & \textbf{61.04} & 81.05 & \textbf{52.22}\\
			\bottomrule  
		\end{tabular}
	}
\end{table*}

\subsection{Implementation Details}
All experiments were conducted on two NVIDIA RTX 4090 GPUs with 24GB of memory each. We trained the model for 50 epochs using the AdamW optimizer with a batch size of 16. The initial learning rate was set to 1e-5 and gradually decayed to 1e-6 by the 30th epoch. The momentum encoder size was set to 65536, and the momentum was set to 0.995. For the input images, we first performed random cropping and horizontal flipping. The cropping region was randomly selected between 50\% and 100\% of the original image, and then uniformly scaled to 256×256 in size. We set the patch size to 4, and the local window size was set to 8. The alpha parameter in the optimized triplet loss and the hyperparameters in the reranking algorithm will be discussed in detail in Section D. For the four stages of the Gswin transformer blocks, we set the third stage to 3 layers, while the remaining stages were set to 1 layer. For the text encoder and multimodal encoder, we employed a total of 12 layers of transformer, with 6 layers for the text encoder and 6 layers for the multimodal encoder. Subsequent experiments also demonstrated that even though the text encoder was smaller than the image encoder, it still achieved excellent performance.
To ensure the stability and reproducibility of our experimental results, we adopted a three random seed strategy, conducting three independent experiments and reporting the average of these results.

\subsection{Method Comparison}

In this section, we compare our approach with the following methods: AMFMN-soft, AMFMN-fusion, AMFMN-sim, LW-MCR-b, LW-MCR-d, GaLR, CLIP, IEFT, HVSA, KAMCL.
\begin{enumerate}
	\item \textit{AMFMN \cite{yuan2022exploring}:} We consider three variants of AMFMN: AMFMN-soft, AMFMN-fusion, and AMFMN-sim. These methods aim to improve remote sensing image retrieval by aligning global and local features, highlighting the integration of global and local information.
	\item \textit{LW-MCR \cite{yuan2021lightweight}:} LW-MCR introduces knowledge distillation to enhance the performance and efficiency of remote sensing image-text retrieval. Depending on the distillation approach, two models are considered: LW-MCR-b employs self-distillation without involving other networks, while LW-MCR-d introduces a teacher network to guide the training of the student network.
	\item \textit{GaLR \cite{yuan2022remote}:} GaLR is a remote sensing image-text retrieval method based on global and local information. It simultaneously considers global and local features to improve retrieval performance, demonstrating the fusion of multi-scale information.
	\item \textit{CLIP \cite{radford2021learning}:}CLIP is a pioneering cross-modal retrieval method that leverages powerful contrastive learning techniques. It is pre-trained on a large-scale image-text dataset, then maps image and text features into a shared space via linear projections. CLIP has demonstrated outstanding performance in numerous cross-modal retrieval tasks, showcasing its broad applicability.
	\item \textit{IEFT \cite{tang2023interacting}:}IEFT is a novel interactive enhanced feature transformer that aims to enhance remote sensing image-text retrieval by interactively improving features. This method explores a novel feature enhancement approach, providing a new perspective for remote sensing image-text retrieval.
	\item \textit{HVSA \cite{zhang2023hypersphere}:}HVSA is a hypersphere-based approach for remote sensing cross-modal text-image retrieval. It introduces an adaptive alignment strategy based on curriculum learning, a feature uniformity strategy, and a key-entity attention mechanism to enhance retrieval performance.
	\item \textit{KAMCL \cite{ji2023knowledge}:}KAMCL is a novel method for remote-sensing image-text retrieval. It introduces a knowledge-aided learning framework, integrates momentum contrastive learning, and employs a hierarchical aggregator module to capture multilevel image information, addressing the challenge of extremely analogous descriptions in RSITR tasks.
\end{enumerate}

Here, CLIP and IEFT models are similar to our approach, both adopting transformer-based architecture. However, other compared models are mainly based on CNN architecture. In our method, part of the layers of the text encoder are reallocated to the multi-modal encoder, so compared to IEFT, our model only adds three layers to the image encoder.

Our proposed CMPAGL method is divided into two variants: CMPAGL without SMR and CMPAGL with SMR, abbreviated as CMPAGL w/o SMR and CMPAGL with SMR, respectively. We conducted multiple experiments on four datasets, and the average results were recorded in Table \ref{table2} to demonstrate the consistency and reproducibility of our method.

\subsubsection{Results on the RSICD Dataset} 
As the largest dataset among the four, RSICD provides a more favorable environment for Transformer-based methods.From the experimental comparisons presented in Table \ref{table2}, it is evident that Transformer-based methods outperform the majority of DCNN-based approaches. This superiority is likely due to the Transformer model's ability to effectively leverage large training datasets, thereby uncovering deeper semantic associations between remote sensing images and texts. Notably, our CMPAGL with SMR method significantly outperforms all other methods across all evaluation metrics. Taking the mR metric as an example, CMPAGL w/o SMR already surpasses all the baselines, and the proposed similarity matrix reweighting (SMR) reranking algorithm further boosts the performance of CMPAGL with SMR by 1.53\%, far exceeding other methods. Specifically, CMPAGL outperforms the AMFMN series by 12.61\%$\sim$13.1\%, the LW-MCR series by 14.11\%$\sim$14.08\%, GaLR by 9.67\%, CLIP by 7.31\%, IEFT by 2.77\%, HVSA by 8.47\%, and KAMCL by 2.53\%.
These exciting results stem from two key innovations: First, CMPAGL pre-aligns image and text features before modal fusion, enabling better fusion and alleviating feature distribution misalignment issues. Second, our designed Gswin transformer block can simultaneously exploit both global and local semantic information in remote sensing images, uncovering rich semantics and improving cross-modal representation quality. The above results validate the effectiveness of our method on the large-scale RSICD dataset.
\subsubsection{Results on the RSITMD Dataset}
We compared the CMPAGL method with other state-of-the-art methods on the RSITMD dataset. The results show that CMPAGL achieves the best performance across all evaluation metrics for both text-to-image and image-to-text retrieval tasks. For the image-to-text retrieval task, taking the R@1 metric as an example, CMPAGL with SMR improves its value from 10.63\%$\sim$11.06\% achieved by the AMFMN series to 19.13\%, outperforming the LW-MCR series by 8.95\%$\sim$10.06\%, GaLR by 4.31\%, CLIP by 4.36\%, IEFT by 3.79\%, HVSA by 5.93\%, and KAMCL by 2.62\%. For the text-to-image retrieval task, CMPAGL also exhibits outstanding performance, improving R@1 from 9.82\%$\sim$11.51\% of the AMFMN series to 15.67\%, outperforming the LW-MCR series by 7.88\%$\sim$9.56\%, GaLR by 4.52\%, CLIP by 5.44\%, IEFT by 4.65\%, HVSA by 4.24\%, and KAMCL by 2.17\%. These remarkable results are attributed to two key factors: First, CMPAGL can learn significant semantic correspondences to address the alignment issue between remote sensing images and texts. Second, our proposed optimized triplet loss can narrow the absolute distance between matching image-text pairs, unlike traditional triplet loss, which only focuses on the relative distance between matching and non-matching pairs.
\subsubsection{Results on the UCM-Captions Dataset} 
Due to the relatively small scale of the UCM-Captions dataset, the results clearly show that DCNN-based methods outperform Transformer-based methods on the R@1 metric. However, benefiting from the transformer model's ability to better capture multi-modal associations between images and texts, Transformer-based methods still exhibit decent performance on comprehensive metrics such as R@5 and R@10. Our CMPAGL with SMR method achieves the best performance on the mR metric. Notably, for the image-to-text retrieval task, CMPAGL with SMR achieves an R@10 of 93.05\%, significantly outperforming all other methods.
\subsubsection{Results on the Sydney-Captions Dataset} 
The Sydney-Captions dataset is the smallest among the four datasets, and the results show that the reduced data scale has some impact on our model's performance. However, thanks to the strong image feature extraction capability of our proposed Gswin transformer block, CMPAGL still outperforms other baselines on the R@1, R@5 metrics for the image-to-text retrieval task, and the mR metric.

\subsection{Ablation Study}

To effectively evaluate the performance of our proposed CMPAGL model, we conducted comprehensive ablation experiments on two publicly available remote sensing image-text matching datasets: RSICD and RSITMD. The ablation study aims to verify the impact of each key module on the final performance, as shown in Table \ref{rsicd} and Table \ref{ablation}. Specifically, we studied the following variants:

\begin{table*}[htbp]
	\caption{Ablation Experiment for Different Components of CMPAGL on RSICD Dataset}
	\label{rsicd}
	\centering
	\begin{tabular}{c|ccccc|c|c|c|c|c|c}
		\toprule
		\multirow{2}{*}{Ablation Model}  & \multirow{2}{*}{GW} & \multirow{2}{*}{SLW}  & \multirow{2}{*}{ITM} & \multirow{2}{*}{O-Tri} & \multirow{2}{*}{SMR} & \multicolumn{3}{c}{Text Retrieval} & \multicolumn{3}{c}{Image Retrieval} \\
		&    & & &  &  & R@1 & R@5 & R@10 & R@1 & R@5 & R@10 \\
		\midrule
		ViT$^{p}$ (baseline)& $\times$ & $\times$ & $\times$ & $\times$&$\times$& 8.62 & 22.78 & 35.15 & 6.18 & 21.51 & 37.17  \\
		
		Gswin$^{p}$ w/o GW & $\times$ &\checkmark & $\times$ &$\times$ & $\times$& 9.24 & 24.07 & 37.47 & 7.32 & 24.96 & 40.22  \\
		
		Gswin$^{p}$ w/o SLW & \checkmark &$\times$  & $\times$ &$\times$ & $\times$& 9.72 & 24.67 & 38.51 & 7.01 & 24.87 & 41.78 \\
		
		Gswin$^{p}$ & \checkmark  & \checkmark  & $\times$ & $\times$ &$\times$ & 10.27 & 25.76 & 39.33 & 7.82 & 26.63 & 42.81  \\
		
		Gswin$^{p}$ + ITM & \checkmark &\checkmark & \checkmark & $\times$ &$\times$ & 10.86 & 26.35 & 40.02 & 7.59 & 27.80 & 43.37  \\
		
		Gswin$^{p}$ + ITM + O-Tri &\checkmark&\checkmark&\checkmark&\checkmark& $\times$& 11.71 & 27.08 & 41.17 & 8.86 & 29.03 & 44.72  \\
		
		Gswin$^{p}$ + ITM + O-Tri + SMR & \checkmark & \checkmark & \checkmark & \checkmark & \checkmark & \textbf{12.71} & \textbf{29.55} & \textbf{42.54} & \textbf{9.08} & \textbf{31.11} & \textbf{46.81} \\
		\bottomrule
	\end{tabular}
\end{table*}

\begin{table*}[htbp]
	\caption{Ablation Experiment for Different Components of CMPAGL on RSITMD Dataset}
	\label{ablation}
	\centering
	\begin{tabular}{c|ccccc|c|c|c|c|c|c}
		\toprule
		\multirow{2}{*}{Ablation Model}  & \multirow{2}{*}{GW} & \multirow{2}{*}{SLW} & \multirow{2}{*}{ITM} & \multirow{2}{*}{O-Tri} & \multirow{2}{*}{SMR} & \multicolumn{3}{c}{Text Retrieval} & \multicolumn{3}{c}{Image Retrieval} \\
		&    & & & &  & R@1 & R@5 & R@10 & R@1 & R@5 & R@10 \\
		\midrule
		ViT$^{p}$ (baseline)& $\times$ & $\times$ & $\times$ & $\times$&$\times$& 11.40 & 28.87 & 42.79 & 9.78 & 33.02 & 46.41  \\
		
		Gswin$^{p}$ w/o GW & $\times$ &\checkmark  & $\times$ &$\times$ & $\times$& 13.87 & 31.82 & 46.21 & 11.25 & 36.57 & 52.81 \\
		
		Gswin$^{p}$ w/o SLW & \checkmark &$\times$  & $\times$ &$\times$ & $\times$& 14.60 & 32.38 & 47.25 & 11.01 & 36.28 & 53.97 \\
		
		Gswin$^{p}$  & \checkmark  & \checkmark &$\times$  & $\times$ &$\times$ & 15.87 & 34.25 & 48.41 & 12.65 & 40.86 & 55.31  \\
		
		Gswin$^{p}$ + ITM & \checkmark & \checkmark & \checkmark & $\times$ &$\times$ & 16.48 & 34.96 & 49.12 & 12.41 & 40.88 & 56.98  \\
		
		Gswin$^{p}$ + ITM + O-Tri & \checkmark&\checkmark&\checkmark&\checkmark& $\times$& 18.81 & 36.72 & 51.11 & 14.48 & 41.65 & 60.35  \\
		
		Gswin$^{p}$ + ITM + O-Tri + SMR & \checkmark  & \checkmark & \checkmark & \checkmark & \checkmark & \textbf{19.13} & \textbf{42.36} & \textbf{54.74} & \textbf{15.67} & \textbf{44.32} & \textbf{60.51} \\
		\bottomrule
	\end{tabular}
\end{table*}

\begin{itemize}
	\item ViT$^{p}$: Uses ViT as the visual encoder with only Image-Text Contrastive (ITC) and Masked Language Modeling (MLM) losses.
	\item Gswin$^{p}$ w/o GW: Uses Gswin without the global window as the visual encoder, where the network only performs window self-attention, with ITC and MLM losses.
	\item Gswin$^{p}$ w/o SLW: Uses Gswin without the shifted local window as the visual encoder, directly performing window attention, with ITC and MLM losses.
	\item Gswin$^{p}$: Uses the complete Gswin transformer as the visual encoder, integrating global and local information, with ITC and MLM losses.
	\item Gswin$^{p}$ + ITM: Adds Image-Text Matching (ITM) loss to the Gswin$^{p}$ model.
	\item Gswin$^{p}$ + ITM + O-Tri: Adds Optimized Triplet loss to the previous configuration.
	\item Gswin$^{p}$ + ITM + O-Tri + SMR: Adds the Similarity Matrix Re-ranking (SMR) algorithm to the previous configuration.
\end{itemize}

The ablation of either the global window (Gswin$^{p}$ w/o GW) or the shifted local window (Gswin$^{p}$ w/o SLW) from the Gswin$^{p}$ architecture results in a substantial degradation of performance across all evaluation metrics. This clearly demonstrates the importance of integrating global and local information through our proposed design, as each component contributes substantially to the overall performance.
The introduction of ITM loss (Gswin$^{p}$ + ITM) further enhances the alignment between images and text, improving cross-modal matching accuracy. 
Notably, the incorporation of ITM led to a reduction in the R@1 metric for image retrieval. This can be attributed to ITM's emphasis on global matching, which may neglect critical fine-grained details necessary for accurately identifying the top-ranked image corresponding to the query. Conversely, the R@1 performance in text retrieval remained unaffected, likely due to the more explicit and structured semantic information in texts. This allows global matching, as influenced by ITM, to be more effective in text retrieval tasks.
The integration of optimized triplet loss (Gswin$^{p}$ + ITM + O-Tri) further enhances the model's capacity to learn discriminative cross-modal representations, substantially diminishing the semantic distance between corresponding image-text pairs. 
This is demonstrated by the substantial improvement in metrics, particularly in the R@1 scores for both text and image retrieval.
The subsequent implementation of SMR algorithm (Gswin$^{p}$ + ITM + O-Tri + SMR) further refines the initial retrieval results, yielding additional enhancements in retrieval precision.
Through the gradual addition of various modules and loss functions, we can clearly observe their incremental impact on model performance. The results of our comprehensive ablation study demonstrate that each architectural component plays a pivotal role in enhancing the overall retrieval performance.

The comparative experiments on the RSICD and RSITMD datasets, as shown in Tables \ref{mi-rsicd} and \ref{mi-rsitmd}, demonstrate that the Pre-Aligned architecture (PA-MI) consistently outperforms the Standard Modal Interaction architecture (Std-MI) across all evaluation metrics. To ensure a fair comparison, the Standard Modal Interaction architecture was divided into two parts: one where the image features were used as queries (Q) to perform modal interaction with the text features, and another where the text features were used as queries to perform modal interaction with the image features. The total number of transformer layers in the Standard Modal Interaction architecture was kept identical to that of the Pre-Aligned architecture.
This setup underscores that performing alignment of image and text features before cross-modal fusion significantly enhances the effectiveness of the fusion process. The results indicate that pre-alignment facilitates better integration of information from both modalities, leading to improved performance in image-text retrieval tasks.

\begin{table}[htbp]
	\centering
	\caption{Comparative Performance of Pre-Aligned and Standard Modal Interaction Architectures on RSICD Dataset for Image-Text Retrieval}
	\label{mi-rsicd}
	\begin{tabular}{ccccccc}
		\toprule
		\multirow{2}{*}{Ablation Model} & \multicolumn{3}{c}{Text Retrieval} & \multicolumn{3}{c}{Image Retrieval} \\
		\cmidrule(r){2-4} \cmidrule(r){5-7}
		& R@1 & R@5 & R@10 & R@1 & R@5 & R@10 \\
		\midrule
		Std-MI & 10.52 & 25.50 & 40.02 & 7.71 & 27.88 & 42.60 \\
		PA-MI & \textbf{11.71} & \textbf{27.08} & \textbf{41.17} & \textbf{8.86} & \textbf{29.03} & \textbf{44.72} \\
		\bottomrule
	\end{tabular}
\end{table}

\begin{table}[htbp]
	\centering
	\caption{Comparative Performance of Pre-Aligned and Standard Modal Interaction Architectures on RSITMD Dataset for Image-Text Retrieval}
	\label{mi-rsitmd}
	\begin{tabular}{ccccccc}
		\toprule
		\multirow{2}{*}{Ablation Model} & \multicolumn{3}{c}{Text Retrieval} & \multicolumn{3}{c}{Image Retrieval} \\
		\cmidrule(r){2-4} \cmidrule(r){5-7}
		& R@1 & R@5 & R@10 & R@1 & R@5 & R@10 \\
		\midrule
		Std-MI & 16.65 & 34.85 & 48.92 & 13.08 & 38.72 & 57.86 \\
		PA-MI & \textbf{18.81} & \textbf{36.72} & \textbf{51.11} & \textbf{14.48} & \textbf{41.65} & \textbf{60.35} \\
		\bottomrule
	\end{tabular}
\end{table}

\begin{table}[htbp]
	\centering
	\caption{Comparative Performance of Different Network Architectures on RSITMD Dataset for Image-Text Retrieval}
	\label{vsbackbone}
	\begin{tabular}{lcccccc}
		\toprule
		\multirow{2}{*}{Method} & \multicolumn{3}{c}{Text Retrieval} & \multicolumn{3}{c}{Image Retrieval} \\
		\cmidrule(r){2-4} \cmidrule(r){5-7}
		& R@1 & R@5 & R@10 & R@1 & R@5 & R@10 \\
		\midrule
		VIT & 12.87 & 31.71 & 45.79 & 10.92 & 35.63 & 52.92 \\
		Swin & 15.85 & 34.27 & 48.83 & 13.72 & 38.99 & 56.14 \\
		GCVIT & 16.72 & 35.61 & 50.15 & 13.81 & 40.39 & 58.68 \\
		Gswin (Ours) & \textbf{18.81} & \textbf{36.72} & \textbf{51.11} & \textbf{14.48} & \textbf{41.65} & \textbf{60.35} \\
		\bottomrule
	\end{tabular}
\end{table}

To further demonstrate the effectiveness of our proposed Gswin transformer architecture, we compared it with several other famous visual encoders, as shown in Table \ref{vsbackbone}:
\begin{itemize}
	\item Uses the standard vit as the visual encoder
	\item Uses Swin transformer as the visual encoder
	\item Uses GCVIT as the visual encoder
\end{itemize}
From the results, we can observe the following points:
The proposed Gswin transformer achieves the best performance among all comparison methods, far superior to ViT, Swin transformer, and GCVIT. This validates the effectiveness of our designed Gswin architecture in capturing global and local visual information of remote sensing images, which is crucial for accurate remote sensing image-text matching.
 
These experiments conclusively validate the effectiveness of the CMPAGL model. The results demonstrate that the pre-alignment of modalities, the Gswin transformer architecture, optimized triplet loss, and SMR each play an essential role in enhancing remote sensing image-text matching on the RSICD and RSITMD dataset.

\subsection{Reranking Parameter Analysis}

In this section, we designed a series of comparative experiments to intuitively demonstrate the advantages of our proposed SMR reranking algorithm. We applied both the SMR and MR reranking algorithms to our CMPAGL and compared the performance of these two algorithms graphically.

Furthermore, we conducted ablation experiments to validate the effectiveness of our proposed extreme difference ratio component and similarity matrix reweighting. We visually demonstrated the efficacy of similarity matrix reweighting by comparing the impact on retrieval metrics with and without it.

As retrieval reranking involves multiple weighted components, determining appropriate weight parameters to balance the contributions of each component is crucial. Therefore, we performed grid search experiments on the RSITMD dataset to explore the impact of different parameter combinations on retrieval performance. The experimental setup is as follows: We selected the average recall rates mR of R@1, R@5, and R@10 for text retrieval and image retrieval to measure performance. For $\gamma_1$ and $\gamma_2$, we set the value range to {0.5, 2.0} with an interval of 0.1, and traversed all possible parameter combinations through grid search.

\begin{figure}[htbp]
	\centering
	\subfloat[]{
		\label{Rerank_Ana_a}
		\includegraphics[width=\linewidth]{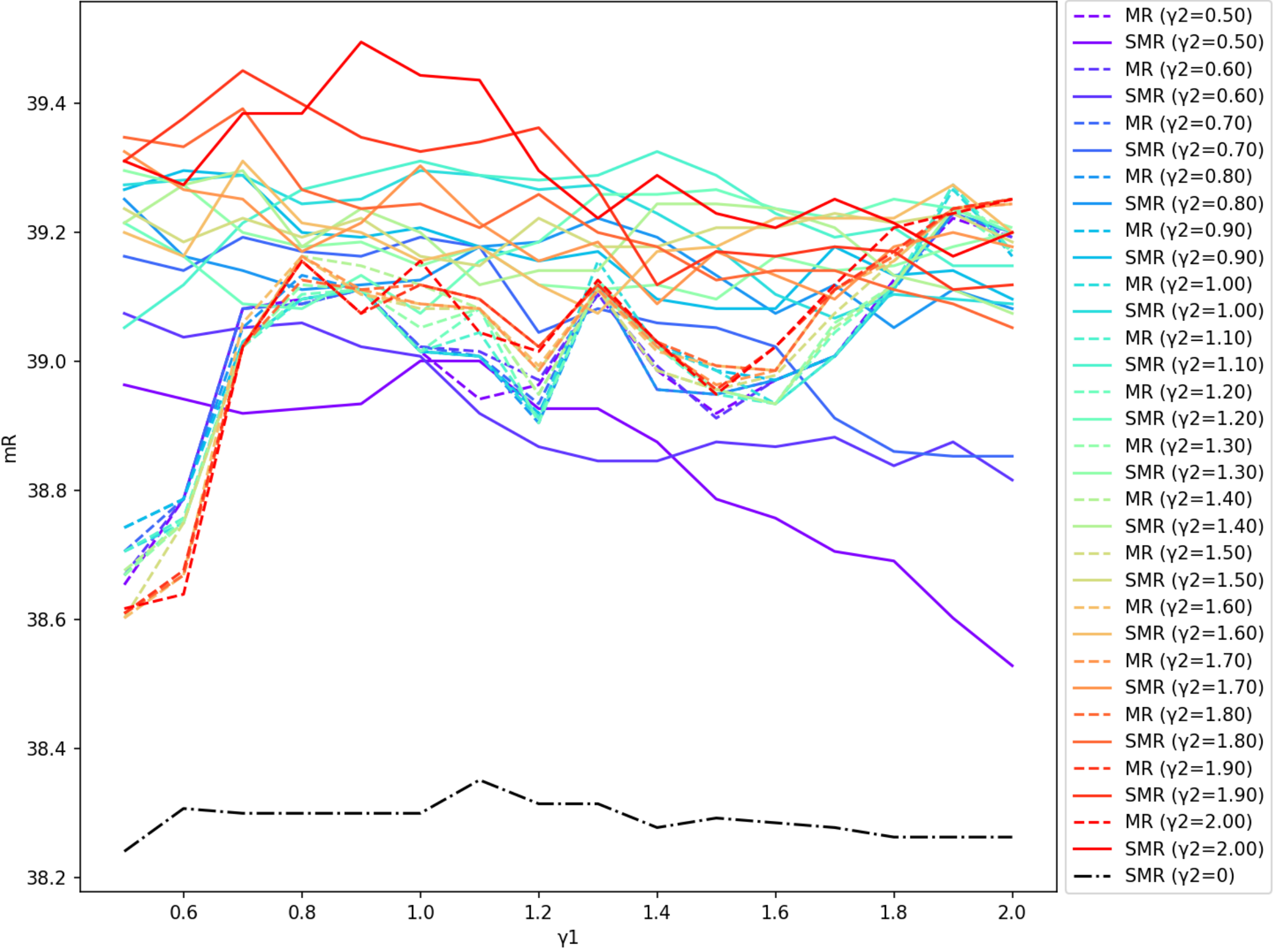}
	}
	
	\subfloat[]{
		\label{Rerank_Ana_b}
		\includegraphics[width=\linewidth]{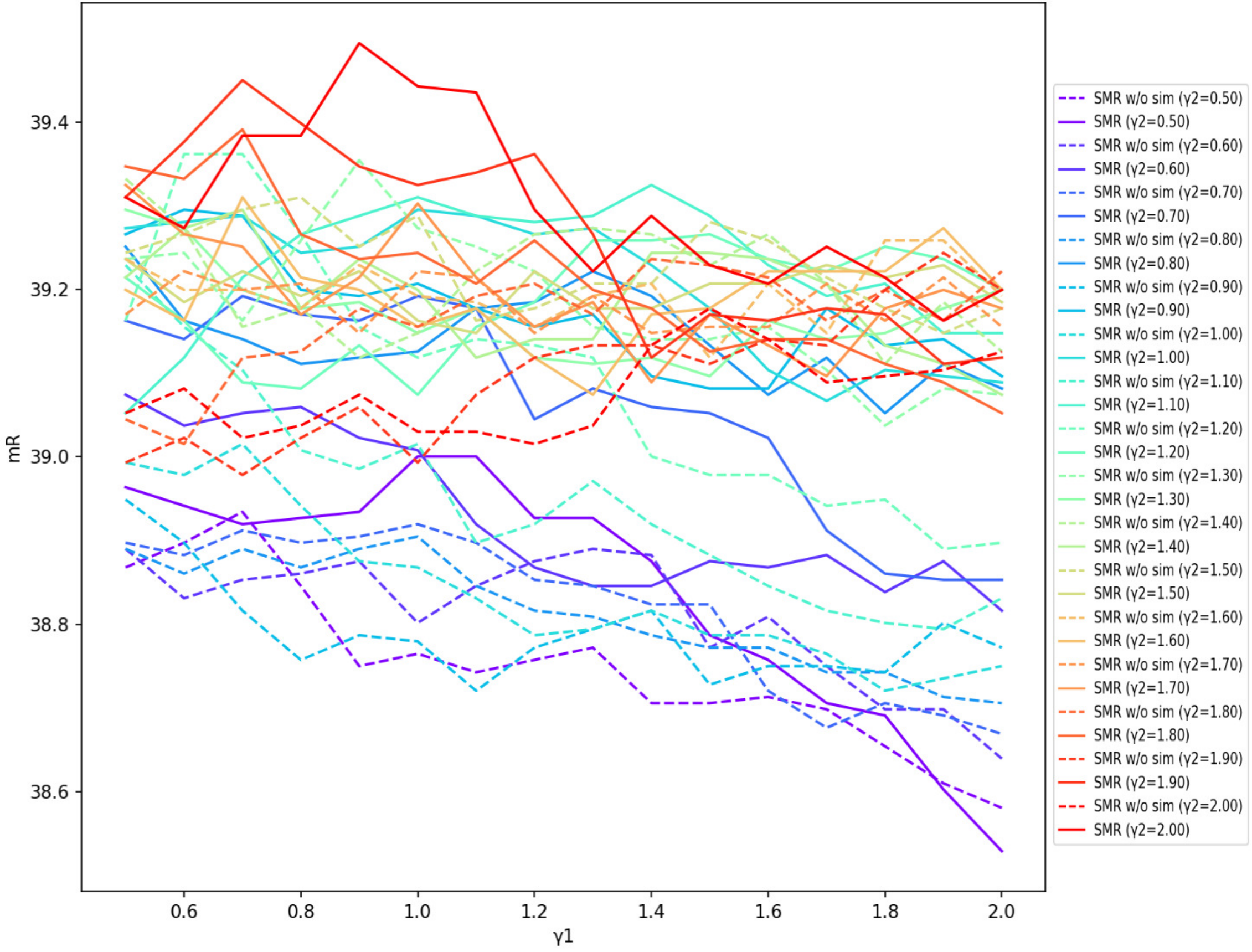}
	}
	\caption{
		Parameter analysis of reranking algorithms: comparative study and ablation experiment. (a) Comparative performance of proposed SMR and MR reranking algorithms under different parameters. (b) Ablation study within SMR: performance comparison with and without re-weighting by original similarity matrix.
	}
	\label{Rerank_Ana}
\end{figure}

The experimental results are presented in line plots in Fig.~\ref{Rerank_Ana}. Fig.~\ref{Rerank_Ana_a} compares the SMR and MR reranking algorithms under the same K value conditions, where the black dashdot line represents SMR without the extreme difference ratio component (i.e., $\gamma_2$ = 0), highlighting the importance of our proposed component. Fig.~\ref{Rerank_Ana_b} presents a comparison between the SMR algorithms with and without the application of similarity matrix reweighting, where 'SMR w/o sim' denotes the SMR algorithm devoid of this reweighting process. From these figures, we can observe the following points:
\begin{itemize}
	\item The significance component of MR is the ratio of the similarity between the query image $i_q$ and the most matched text to the sum of similarities between $i_q$ and all texts. Therefore, the variation of $\gamma_2$ has little overall impact, resulting in curves with almost the same shape for different $\gamma_2$ values. In contrast, our extreme difference ratio component primarily reflects the difference between the similarity of the current query image $i_q$ and the retrieved text $t_j$, and the similarity between $i_q$ and the most matched text. Thus, it is more sensitive to the variation of $\gamma_2$, exhibiting significant changes under different $\gamma_2$ states.
	
	\item In most parameter intervals, the maximum mR value of the SMR algorithm is higher than or at least equal to the highest point of the MR algorithm, reflecting the higher degree of freedom in parameter selection for SMR and its ability to achieve optimal performance more easily.
	
	\item The SMR method achieves the best performance when $\gamma_1$ = 0.9 and $\gamma_2$ = 1.9.
	\item The bottom dashdot line represents SMR without the extreme difference ratio component (i.e., $\gamma_2$ = 0), which is significantly lower than SMR with the proposed component, strongly validating the effectiveness of our extreme difference ratio component.
	\item From Fig.~\ref{Rerank_Ana_b}, it is evident that the SMR algorithm, endowed with the similarity matrix reweighting, exhibits superior performance.
\end{itemize}

\begin{figure*}[!t]
	\centering
	\includegraphics[width=\linewidth]{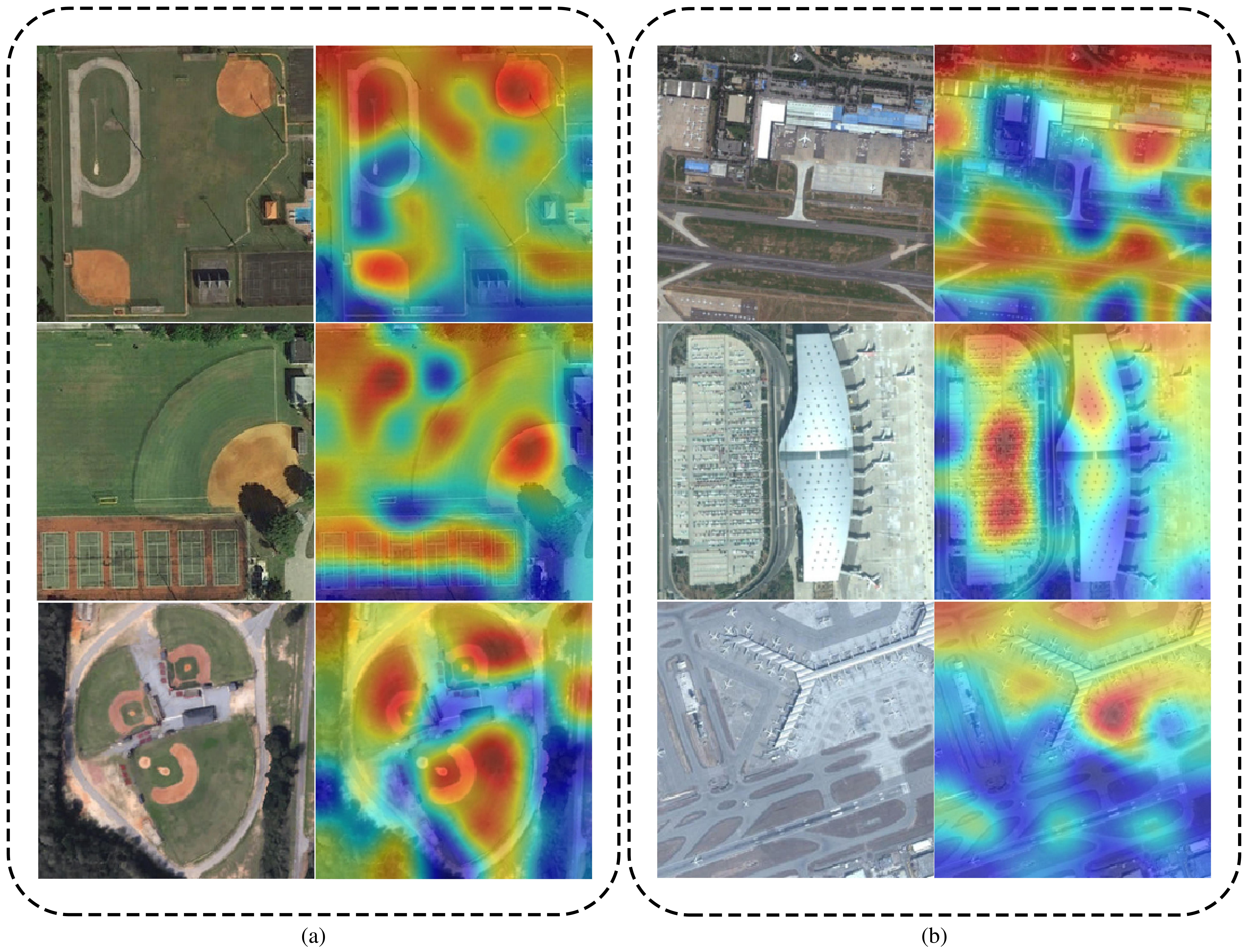}
	\caption{
		Visualization of image encoder output in cross-modal retrieval network. (a)Visualization under baseball field scene. (b)Visualization under airport scene.
	}
	\label{heatmap}
\end{figure*}

\begin{figure}[htbp]
	\centering
	\subfloat[]{
		\label{retrieval_vis_1}
		\includegraphics[width=\linewidth]{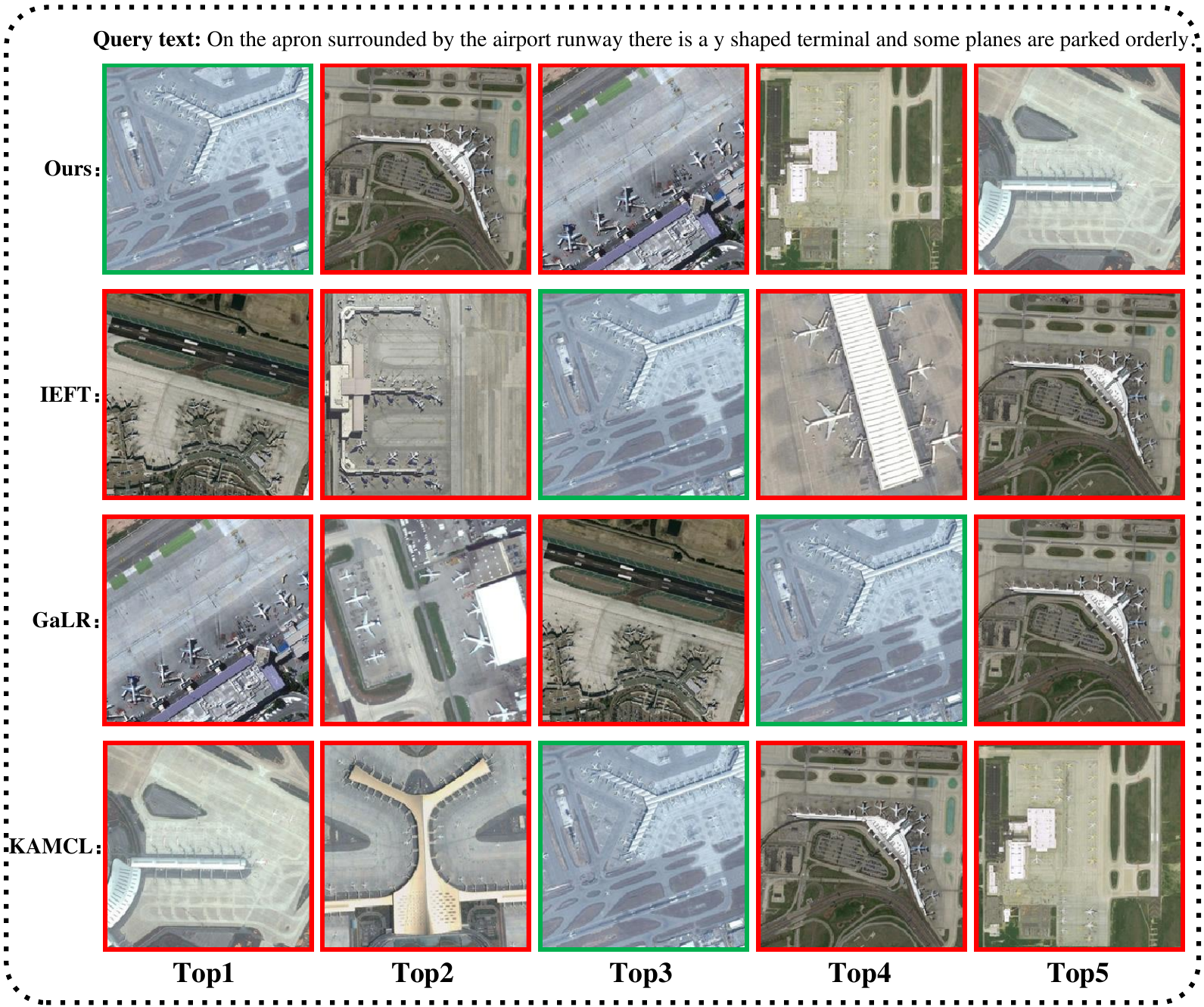}
	}
	
	\subfloat[]{
		\label{retrieval_vis_2}
		\includegraphics[width=\linewidth]{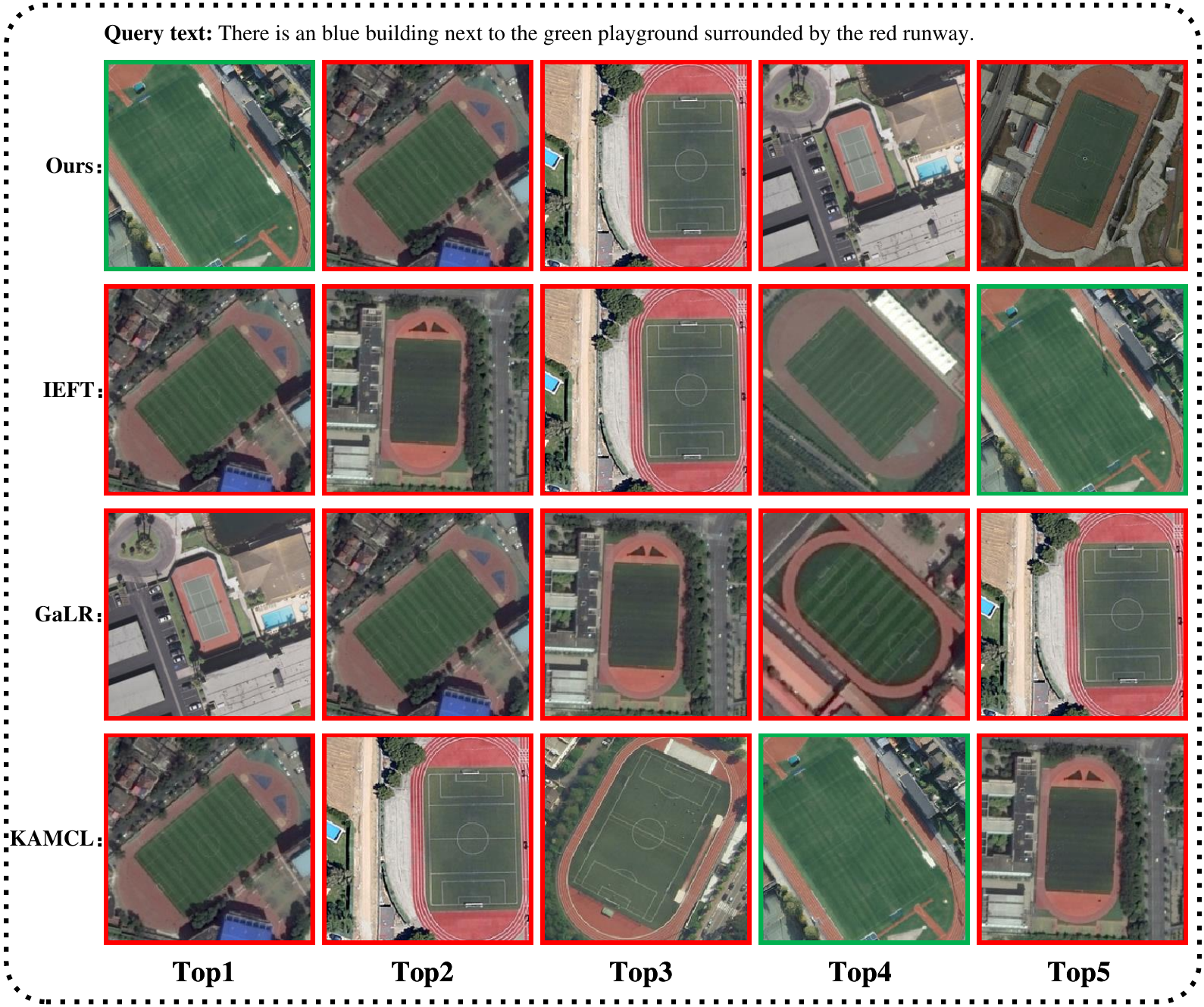}
	}
	\caption{Comparison of query text and retrieved images by different methods. Matching images are highlighted with a green frame, while non-matching images are highlighted with a red frame.
		(a) Retrieving images matching the described Y-shaped terminal and parked planes on the apron surrounding the airport runway using the query text,comparing the retrieval results of different methods.
		(b) Retrieving images matching the described blue building, green playground, and red runway using the query text, comparing the retrieval results of different methods.
	}
	\label{retrieval_vis}
\end{figure}

\subsection{Attention Map Analysis for Retrieval Performance}

Fig.~\ref{heatmap} presents the attention maps for the airport and baseball field scenes. The Gswin transformer block, which is central to our proposed method, is instrumental in generating these visualizations. The left side of each panel shows the original satellite image, while the right side overlays the image with a heatmap that visualizes the attention weights derived from the last layer of the Gswin transformer block. These attention weights reflect the intricate fusion of global and local information, as captured by our Gswin's innovative architecture.
In Fig.~\ref{heatmap}(a), the heatmap for the first image reveals a strong focus on the areas described in the text, such as ``Two baseball fields and several tennis courts and a track and field," which highlights the Gswin transformer's ability to effectively capture and fuse semantic details across different scales. This is a direct outcome of the Gswin block's dual-branch structure, where the global-local window attention mechanism ensures that both fine-grained local features and broader global context are synergistically integrated, thereby enhancing the model’s scene comprehension.
The attention maps for the second and third images in Fig.~\ref{heatmap}(a) further demonstrate the Gswin block's proficiency in concentrating on areas critical to scene semantics, underscoring the superiority of our model in understanding complex scene layouts.

In Fig.~\ref{heatmap}(b), the visual model’s ability to extract local features, particularly small targets like cars and planes, is put to the test. The Gswin transformer block effectively highlights these small yet significant elements, such as the planes on the left side of the first image, which are critical for accurate scene understanding. Additionally, the Y-shaped building’s heatmap in the third image effectively aligns with the text description, demonstrating the model’s precise alignment between image features and text descriptions, which is supported by our optimized triplet loss. This loss function not only ensures tighter clustering of matching image-text pairs but also minimizes intra-class distances, contributing to more accurate cross-modal retrievals.

Overall, the continuous and concentrated attention patterns observed in both airport and baseball field scenes underscore the robustness of our proposed model. By leveraging the Gswin transformer's unique architectural innovations and the optimized triplet loss, our model not only accurately identifies key elements within scenes but also allocates attention weights in a manner that highlights essential features, contributing to enhanced generalization across different types of scenes. These results demonstrates the effectiveness of the effectiveness of our cross-modal model architecture in remote sensing image understanding and retrieval.

\subsection{Analysis of Retrieval Results}

In this section, we conduct a visual analysis of the retrieval results compared to the IEFT, GaLR, and KAMCL methods, employing a text-to-image retrieval paradigm and examining the top 5 retrieved images. Correctly retrieved images are highlighted with green bounding boxes to clearly demonstrate the advantages of our model. It's worth noting that none of the retrieval results employed reranking algorithms.

As illustrated in Fig.~\ref{retrieval_vis_1}, the query text describes: ``On the apron surrounded by the airport runway, there is a Y-shaped terminal, and some planes are parked orderly." While the plane targets are relatively small, establishing their connection with the large Y-shaped terminal mentioned in the description poses a significant challenge. The crux lies in establishing a semantic link between these small targets (planes) and the described Y-shaped terminal to achieve more accurate matching. By analyzing the results, we observe:

All four methods successfully retrieved images matching the query text, but the IEFT method retrieved the correct image at the third position, the GaLR method retrieved it at the fourth position, and the KAMCL method also at the third position. Our method accurately retrieved the image that most closely matched the query text. Examining the retrieved images from different methods, we can see that they all managed to retrieve the runway, planes, and terminal buildings mentioned in the text. However, the precise description of the Y-shaped terminal presented a significant challenge for the models' cross-modal interactions. The IEFT method, which directly feeds the image and text into a transformer network for cross-modal interaction, struggles to learn the deep semantic relationships between images and text on this small dataset. The GaLR method lacks a cross-modal interaction process and has high demands on object detection accuracy, making it difficult to capture deep semantic relationships. The KAMCL method, while better at distinguishing subtle differences through its knowledge-aided learning framework, still falls short in accurately capturing the complex spatial relationships in this particular scenario. In contrast, our method effectively retrieved multiple highly relevant images that accurately depict the described airport scene, including the Y-shaped terminal and parked planes, accurately matching the query text.

In the second example Fig.~\ref{retrieval_vis_2}, the query text describes: ``There is a blue building next to the green playground surrounded by the red runway." Similarly, the green-bordered images are considered the most relevant matches. From these results, we can observe:

By examining the retrieved most relevant images, we can see that the small blue target is easily overlooked, while the playground and runway are large-scale targets, posing challenges for the network's local feature extraction and the fusion of local and global features.
Our method successfully retrieved multiple highly relevant images that accurately depict the described scene, including the blue building, green playground, and red runway. This is attributable to our proposed Gswin transformer, which fuses the global features extracted from the global window with local features, enabling the establishment of relationships between the green playground and blue building, and better attending to small targets. The KAMCL method, benefiting from its hierarchical aggregator (HA) mechanism, which captures multilevel and comprehensive visual representations, retrieved the correct matching image at the fourth position. The IEFT method retrieved the correct matching image at the fifth position, while the GaLR method did not retrieve any satisfactory matches. Although the IEFT and GaLR methods retrieved some relatively relevant images containing blue buildings, green playgrounds, and red runways, the spatial relationships were not entirely accurate.

Overall, the above results powerfully demonstrate the superior performance of our proposed method in retrieving images that are highly relevant to the given query text descriptions. By effectively fusing the semantic features of images and text, our model can accurately capture the key visual elements described in the query text, significantly outperforming existing techniques in terms of matching accuracy. This outstanding performance highlights the effectiveness of our proposed cross-modal pre-alignment strategy, enabling the model to establish closer semantic associations between images and text even before the cross-modal interaction, laying a solid foundation for the subsequent cross-modal fusion process. 

\begin{table}[htbp]
	\centering
	\caption{Comparison of Evaluation Times, Inference Times, and Parameter Counts for Different Methods across RSICD, RSITMD, and UCM Datasets}
	\label{tab:times}
	\begin{tabular}{c|ccc|c|c}
		\toprule 
		\multirow{2}{*}{Method} & \multicolumn{3}{c|}{Evaluation Time(s)} & \multirow{2}{*}{Inference Time(s)} & \multirow{2}{*}{Params(M)}\\
		& RSICD & RSITMD & UCM & &\\
		\midrule
		AMFMN  & 65.23  & 15.14  & 8.47  & 0.0711 & 36.70\\
		LW-MCR & 28.12  & 7.36   & 3.27  & 0.0093 & 1.65\\
		GaLR   & 67.07  & 16.69  & 9.03  & 0.0710 & 46.89\\
		CLIP   & 402.32 & 102.09 & 36.29 & 0.6175 & 151.35\\
		IEFT   & 328.17 & 75.39  & 27.66 & 0.4137 & 100.12\\
		HVSA   & 48.15  & 13.67  & 8.04  & 0.0597 & 35.01\\		
		KAMCL  & 44.27  & 12.47  & 7.87  & 0.0502 & 40.86\\
		CMPAGL & 191.01 & 41.55  & 16.87 & 0.1673 & 121.21 (89.35)\\
		\bottomrule 
	\end{tabular}
\end{table}

\subsection{Discussion on Time Cost and Model Parameters}

This section presents a comprehensive analysis of model parameter count, evaluation time, and inference time, comparing our model with other excellent methods.
Evaluation time is defined as the total duration required to compute similarities between all image-text pairs across various test sets, serving as a metric for the model's efficiency in processing large-scale datasets.
Inference time denotes the time required for a single cross-modal similarity computation, providing a crucial indicator of the model's responsiveness in individual tasks. 
To enhance result reliability, multiple experiments were conducted, with the averaged outcomes presented in Table \ref{tab:times}.

In Table \ref{tab:times}, the total parameter count for our CMPAGL model is listed as 121.21M, while the value in parentheses (89.35M) represents the reduced parameter count during inference, when the multi-modal encoder is not involved in the computation.
Table \ref{tab:times} illustrates that transformer-based architectures, including CLIP, IEFT, and our proposed CMPAGL, exhibit substantially higher parameter counts relative to alternative models. 
While these large-scale parameters provide powerful feature extraction and cross-modal understanding capabilities, they also result in longer evaluation and inference times.
Although our CMPAGL model has a relatively high parameter count, during the inference stage, cross-modal similarity can be calculated directly after feature extraction by the image and text encoders, without requiring the participation of a multi-modal encoder in the computation. This design allows CMPAGL to achieve more balanced evaluation and inference times while maintaining a high parameter count, as the parameter count is reduced to 89.35M during inference.
Consequently, CMPAGL maintains superior performance in cross-modal similarity computations while concurrently optimizing time efficiency, thus achieving an advantageous equilibrium between performance and computational efficiency.
Conversely, lightweight architectures like LW-MCR, comprising merely 1.65M parameters, exhibit exceptional time efficiency, particularly in inference scenarios. 
Nevertheless, the limited parameter capacity of these models often results in suboptimal performance when confronted with complex tasks, particularly in the domain of intricate remote sensing image description.
Compared to methods such as AMFMN, GaLR, HVSA, and KAMCL, CMPAGL may show slight disadvantages in evaluation and inference times, but it achieves significant improvements in performance. Specifically, CMPAGL outperforms these methods on key evaluation metrics (such as R@1, R@5, R@10, etc.), demonstrating stronger cross-modal retrieval capabilities. This performance advantage is crucial in handling complex remote sensing image description tasks, indicating that CMPAGL has achieved superior feature representation capabilities by reasonably balancing time efficiency and performance, making it a model choice that strikes a good balance between performance and time.

\section{Conclusion}
\label{Conclusion}
Addressing large-scale target variations in remote sensing images, we propose an innovative cross-modal retrieval method with a pre-alignment strategy for bidirectional remote sensing image-text retrieval. We designed a Gswin transformer block with global-local window interaction, generating global windows for cross-attention with local windows to efficiently fuse multi-scale visual features. This captures semantic target information in local windows while enhancing adaptability to small targets and complex scenes.

We adopted Bert with a masked language model (MLM) as the text encoder, leveraging MLM to enhance semantic text-image associations through word prediction from image information. Pre-aligning extracted text and image features before cross-modal interaction reduces fusion difficulty and improves matching accuracy, utilizing internal modal associations for effective cross-modal grounding.

Further, we optimized the triplet loss with an intra-class term to bring matching pairs closer. We proposed a similarity matrix reweighting reranking algorithm with an extreme difference ratio component to exploit information in the original similarity matrix, reweighting it by fusing ranking probability and this component.

These innovations enable outstanding performance on remote sensing image-text retrieval benchmarks, significantly outperforming prior techniques and highlighting the effectiveness of modal pre-alignment, multi-scale fusion, and reranking strategies. While remarkable, we will explore more efficient attention for enhanced small target and complex scene processing. We believe continuous innovation in cross-modal retrieval will powerfully support remote sensing image understanding and mining.

\bibliography{mybibfile}

\begin{thebibliography}{10}

\bibitem{ma2015remote}
Y.~Ma, H.~Wu, L.~Wang, B.~Huang, R.~Ranjan, A.~Zomaya, and W.~Jie, ``Remote
  sensing big data computing: Challenges and opportunities,'' {\em Future
  Generation Computer Systems}, vol.~51, pp.~47--60, 2015.

\bibitem{chi2016big}
M.~Chi, A.~Plaza, J.~A. Benediktsson, Z.~Sun, J.~Shen, and Y.~Zhu, ``Big data
  for remote sensing: Challenges and opportunities,'' {\em Proceedings of the
  IEEE}, vol.~104, no.~11, pp.~2207--2219, 2016.

\bibitem{srivastava2019understanding}
S.~Srivastava, J.~E. Vargas-Munoz, and D.~Tuia, ``Understanding urban landuse
  from the above and ground perspectives: A deep learning, multimodal
  solution,'' {\em Remote Sensing of Environment}, vol.~228, pp.~129--143,
  2019.

\bibitem{borana2019hyperspectral}
S.~Borana, S.~Yadav, and S.~Parihar, ``Hyperspectral data analysis for arid
  vegetation species: Smart \& sustainable growth,'' in {\em 2019 International
  Conference on Computing, Communication, and Intelligent Systems (ICCCIS)},
  pp.~495--500, IEEE, 2019.

\bibitem{amani2020google}
M.~Amani, A.~Ghorbanian, S.~A. Ahmadi, M.~Kakooei, A.~Moghimi, S.~M.
  Mirmazloumi, S.~H.~A. Moghaddam, S.~Mahdavi, M.~Ghahremanloo, S.~Parsian,
  {\em et~al.}, ``Google earth engine cloud computing platform for remote
  sensing big data applications: A comprehensive review,'' {\em IEEE Journal of
  Selected Topics in Applied Earth Observations and Remote Sensing}, vol.~13,
  pp.~5326--5350, 2020.

\bibitem{cheng2022nwpu}
Q.~Cheng, H.~Huang, Y.~Xu, Y.~Zhou, H.~Li, and Z.~Wang, ``Nwpu-captions dataset
  and mlca-net for remote sensing image captioning,'' {\em IEEE Transactions on
  Geoscience and Remote Sensing}, vol.~60, pp.~1--19, 2022.

\bibitem{abdullah2020textrs}
T.~Abdullah, Y.~Bazi, M.~M. Al~Rahhal, M.~L. Mekhalfi, L.~Rangarajan, and
  M.~Zuair, ``Textrs: Deep bidirectional triplet network for matching text to
  remote sensing images,'' {\em Remote Sensing}, vol.~12, no.~3, p.~405, 2020.

\bibitem{cheng2021deep}
Q.~Cheng, Y.~Zhou, P.~Fu, Y.~Xu, and L.~Zhang, ``A deep semantic alignment
  network for the cross-modal image-text retrieval in remote sensing,'' {\em
  IEEE Journal of Selected Topics in Applied Earth Observations and Remote
  Sensing}, vol.~14, pp.~4284--4297, 2021.

\bibitem{lv2021fusion}
Y.~Lv, W.~Xiong, X.~Zhang, and Y.~Cui, ``Fusion-based correlation learning
  model for cross-modal remote sensing image retrieval,'' {\em IEEE Geoscience
  and Remote Sensing Letters}, vol.~19, pp.~1--5, 2021.

\bibitem{sippel2023cross}
F.~Sippel, J.~Seiler, and A.~Kaup, ``Cross spectral image reconstruction using
  a deep guided neural network,'' in {\em 2023 IEEE International Conference on
  Image Processing (ICIP)}, pp.~226--230, IEEE, 2023.

\bibitem{grosche2023image}
S.~Grosche, A.~Regensky, J.~Seiler, and A.~Kaup, ``Image super-resolution using
  t-tetromino pixels,'' in {\em Proceedings of the IEEE/CVF Conference on
  Computer Vision and Pattern Recognition}, pp.~9989--9998, 2023.

\bibitem{yuan2022exploring}
Z.~Yuan, W.~Zhang, K.~Fu, X.~Li, C.~Deng, H.~Wang, and X.~Sun, ``Exploring a
  fine-grained multiscale method for cross-modal remote sensing image
  retrieval,'' {\em IEEE Transactions on Geoscience and Remote Sensing},
  vol.~60, pp.~1--19, 2022.

\bibitem{yuan2022remote}
Z.~Yuan, W.~Zhang, C.~Tian, X.~Rong, Z.~Zhang, H.~Wang, K.~Fu, and X.~Sun,
  ``Remote sensing cross-modal text-image retrieval based on global and local
  information,'' {\em IEEE Transactions on Geoscience and Remote Sensing},
  vol.~60, pp.~1--16, 2022.

\bibitem{yuan2021lightweight}
Z.~Yuan, W.~Zhang, X.~Rong, X.~Li, J.~Chen, H.~Wang, K.~Fu, and X.~Sun, ``A
  lightweight multi-scale crossmodal text-image retrieval method in remote
  sensing,'' {\em IEEE Transactions on Geoscience and Remote Sensing}, vol.~60,
  pp.~1--19, 2021.

\bibitem{zhang2023exploring}
S.~Zhang, Y.~Li, and S.~Mei, ``Exploring uni-modal feature learning on entities
  and relations for remote sensing cross-modal text-image retrieval,'' {\em
  IEEE Transactions on Geoscience and Remote Sensing}, vol.~61, pp.~1--17,
  2023.

\bibitem{zhang2023hypersphere}
W.~Zhang, J.~Li, S.~Li, J.~Chen, W.~Zhang, X.~Gao, and X.~Sun,
  ``Hypersphere-based remote sensing cross-modal text-image retrieval via
  curriculum learning,'' {\em IEEE Transactions on Geoscience and Remote
  Sensing}, 2023.

\bibitem{dosovitskiy2020image}
A.~Dosovitskiy, L.~Beyer, A.~Kolesnikov, D.~Weissenborn, X.~Zhai,
  T.~Unterthiner, M.~Dehghani, M.~Minderer, G.~Heigold, S.~Gelly, J.~Uszkoreit,
  and N.~Houlsby, ``An image is worth 16x16 words: Transformers for image
  recognition at scale,'' {\em ArXiv}, vol.~abs/2010.11929, 2020.

\bibitem{sak2014long}
H.~Sak, A.~W. Senior, and F.~Beaufays, ``Long short-term memory recurrent
  neural network architectures for large scale acoustic modeling,'' in {\em
  Interspeech}, 2014.

\bibitem{chung2014empirical}
J.~Chung, Çaglar G{\"u}lçehre, K.~Cho, and Y.~Bengio, ``Empirical evaluation
  of gated recurrent neural networks on sequence modeling,'' {\em ArXiv},
  vol.~abs/1412.3555, 2014.

\bibitem{vaswani2017attention}
A.~Vaswani, N.~Shazeer, N.~Parmar, J.~Uszkoreit, L.~Jones, A.~N. Gomez,
  {\L}.~Kaiser, and I.~Polosukhin, ``Attention is all you need,'' {\em Advances
  in Neural Information Processing Systems}, vol.~30, 2017.

\bibitem{chen2023multiscale}
Y.~Chen, J.~Huang, X.~Li, S.~Xiong, and X.~Lu, ``Multiscale salient alignment
  learning for remote sensing image-text retrieval,'' {\em IEEE Transactions on
  Geoscience and Remote Sensing}, 2023.

\bibitem{tang2023interacting}
X.~Tang, Y.~Wang, J.~Ma, X.~Zhang, F.~Liu, and L.~Jiao, ``Interacting-enhancing
  feature transformer for cross-modal remote sensing image and text
  retrieval,'' {\em IEEE Transactions on Geoscience and Remote Sensing}, 2023.

\bibitem{li2021align}
J.~Li, R.~Selvaraju, A.~Gotmare, S.~Joty, C.~Xiong, and S.~C.~H. Hoi, ``Align
  before fuse: Vision and language representation learning with momentum
  distillation,'' in {\em Advances in Neural Information Processing Systems}
  (M.~Ranzato, A.~Beygelzimer, Y.~Dauphin, P.~Liang, and J.~W. Vaughan, eds.),
  vol.~34, pp.~9694--9705, Curran Associates, Inc., 2021.

\bibitem{chen2023deep}
Y.~Chen, J.~Huang, L.~Mou, P.~Jin, S.~Xiong, and X.~X. Zhu, ``Deep saliency
  smoothing hashing for drone image retrieval,'' {\em IEEE Transactions on
  Geoscience and Remote Sensing}, vol.~61, pp.~1--13, 2023.

\bibitem{zhao2023multitask}
M.~Zhao, X.~Zhang, and A.~Kaup, ``Multitask learning for sar ship detection
  with gaussian-mask joint segmentation,'' {\em IEEE Transactions on Geoscience
  and Remote Sensing}, 2023.

\bibitem{liu2021swin}
Z.~Liu, Y.~Lin, Y.~Cao, H.~Hu, Y.~Wei, Z.~Zhang, S.~Lin, and B.~Guo, ``Swin
  transformer: Hierarchical vision transformer using shifted windows,'' in {\em
  Proceedings of the IEEE/CVF International Conference on Computer Vision},
  pp.~10012--10022, 2021.

\bibitem{hatamizadeh2023global}
A.~Hatamizadeh, H.~Yin, G.~Heinrich, J.~Kautz, and P.~Molchanov, ``Global
  context vision transformers,'' in {\em International Conference on Machine
  Learning}, pp.~12633--12646, PMLR, 2023.

\bibitem{wang2019matching}
T.~Wang, X.~Xu, Y.~Yang, A.~Hanjalic, H.~T. Shen, and J.~Song, ``Matching
  images and text with multi-modal tensor fusion and re-ranking,'' in {\em
  Proceedings of the 27th ACM International Conference on Multimedia},
  pp.~12--20, 2019.

\bibitem{faghri2017vse++}
F.~Faghri, D.~J. Fleet, J.~R. Kiros, and S.~Fidler, ``Vse++: Improving
  visual-semantic embeddings with hard negatives,'' in {\em British Machine
  Vision Conference}, 2017.

\bibitem{lu2017exploring}
X.~Lu, B.~Wang, X.~Zheng, and X.~Li, ``Exploring models and data for remote
  sensing image caption generation,'' {\em IEEE Transactions on Geoscience and
  Remote Sensing}, vol.~56, no.~4, pp.~2183--2195, 2017.

\bibitem{qu2016deep}
B.~Qu, X.~Li, D.~Tao, and X.~Lu, ``Deep semantic understanding of high
  resolution remote sensing image,'' in {\em 2016 International Conference on
  Computer, Information and Telecommunication Systems (Cits)}, pp.~1--5, IEEE,
  2016.

\bibitem{you2018end}
Q.~You, Z.~Zhang, and J.~Luo, ``End-to-end convolutional semantic embeddings,''
  in {\em Proceedings of the IEEE Conference on Computer Vision and Pattern
  Recognition}, pp.~5735--5744, 2018.

\bibitem{wang2021cross}
C.~Wang, L.~Li, C.~Yan, Z.~Wang, Y.~Sun, and J.~Zhang, ``Cross-modal semantic
  correlation learning by bi-cnn network,'' {\em IET Image Processing},
  vol.~15, no.~14, pp.~3674--3684, 2021.

\bibitem{zheng2020dual}
Z.~Zheng, L.~Zheng, M.~Garrett, Y.~Yang, M.~Xu, and Y.-D. Shen, ``Dual-path
  convolutional image-text embeddings with instance loss,'' {\em ACM
  Transactions on Multimedia Computing, Communications, and Applications
  (TOMM)}, vol.~16, no.~2, pp.~1--23, 2020.

\bibitem{zhen2019deep}
L.~Zhen, P.~Hu, X.~Wang, and D.~Peng, ``Deep supervised cross-modal
  retrieval,'' in {\em Proceedings of the IEEE/CVF Conference on Computer
  Vision and Pattern Recognition}, pp.~10394--10403, 2019.

\bibitem{huang2018learning}
Y.~Huang, Q.~Wu, C.~Song, and L.~Wang, ``Learning semantic concepts and order
  for image and sentence matching,'' in {\em Proceedings of the IEEE Conference
  on Computer Vision and Pattern Recognition}, pp.~6163--6171, 2018.

\bibitem{lee2018stacked}
K.-H. Lee, X.~Chen, G.~Hua, H.~Hu, and X.~He, ``Stacked cross attention for
  image-text matching,'' in {\em Proceedings of the European Conference on
  Computer Vision (ECCV)}, pp.~201--216, 2018.

\bibitem{xu2020cross}
X.~Xu, T.~Wang, Y.~Yang, L.~Zuo, F.~Shen, and H.~T. Shen, ``Cross-modal
  attention with semantic consistence for image--text matching,'' {\em IEEE
  Transactions on Neural Networks and Learning Systems}, vol.~31, no.~12,
  pp.~5412--5425, 2020.

\bibitem{li2019visual}
K.~Li, Y.~Zhang, K.~Li, Y.~Li, and Y.~Fu, ``Visual semantic reasoning for
  image-text matching,'' in {\em Proceedings of the IEEE/CVF International
  Conference on Computer Vision}, pp.~4654--4662, 2019.

\bibitem{li2022image}
K.~Li, Y.~Zhang, K.~Li, Y.~Li, and Y.~Fu, ``Image-text embedding learning via
  visual and textual semantic reasoning,'' {\em IEEE Transactions on Pattern
  Analysis and Machine Intelligence}, vol.~45, no.~1, pp.~641--656, 2022.

\bibitem{lu2019vilbert}
J.~Lu, D.~Batra, D.~Parikh, and S.~Lee, ``Vilbert: Pretraining task-agnostic
  visiolinguistic representations for vision-and-language tasks,'' {\em
  Advances in Neural Information Processing Systems}, vol.~32, 2019.

\bibitem{tan2019lxmert}
H.~H. Tan and M.~Bansal, ``Lxmert: Learning cross-modality encoder
  representations from transformers,'' in {\em Conference on Empirical Methods
  in Natural Language Processing}, 2019.

\bibitem{gao2020fashionbert}
D.~Gao, L.~Jin, B.~Chen, M.~Qiu, P.~Li, Y.~Wei, Y.~Hu, and H.~Wang,
  ``Fashionbert: Text and image matching with adaptive loss for cross-modal
  retrieval,'' in {\em Proceedings of the 43rd International ACM SIGIR
  Conference on Research and Development in Information Retrieval},
  pp.~2251--2260, 2020.

\bibitem{chen2021learning}
J.~Chen, H.~Hu, H.~Wu, Y.~Jiang, and C.~Wang, ``Learning the best pooling
  strategy for visual semantic embedding,'' in {\em Proceedings of the IEEE/CVF
  Conference on Computer Vision and Pattern Recognition}, pp.~15789--15798,
  2021.

\bibitem{huang2020pixel}
Z.~Huang, Z.~Zeng, B.~Liu, D.~Fu, and J.~Fu, ``Pixel-bert: Aligning image
  pixels with text by deep multi-modal transformers,'' {\em ArXiv},
  vol.~abs/2004.00849, 2020.

\bibitem{radford2021learning}
A.~Radford, J.~W. Kim, C.~Hallacy, A.~Ramesh, G.~Goh, S.~Agarwal, G.~Sastry,
  A.~Askell, P.~Mishkin, J.~Clark, {\em et~al.}, ``Learning transferable visual
  models from natural language supervision,'' in {\em International Conference
  on Machine Learning}, pp.~8748--8763, PMLR, 2021.

\bibitem{cheng2022vista}
M.~Cheng, Y.~Sun, L.~Wang, X.~Zhu, K.~Yao, J.~Chen, G.~Song, J.~Han, J.~Liu,
  E.~Ding, {\em et~al.}, ``Vista: Vision and scene text aggregation for
  cross-modal retrieval,'' in {\em Proceedings of the IEEE/CVF Conference on
  Computer Vision and Pattern Recognition}, pp.~5184--5193, 2022.

\bibitem{kim2021vilt}
W.~Kim, B.~Son, and I.~Kim, ``Vilt: Vision-and-language transformer without
  convolution or region supervision,'' in {\em International Conference on
  Machine Learning}, pp.~5583--5594, PMLR, 2021.

\bibitem{bao2022vlmo}
H.~Bao, W.~Wang, L.~Dong, Q.~Liu, O.~K. Mohammed, K.~Aggarwal, S.~Som, S.~Piao,
  and F.~Wei, ``Vlmo: Unified vision-language pre-training with
  mixture-of-modality-experts,'' {\em Advances in Neural Information Processing
  Systems}, vol.~35, pp.~32897--32912, 2022.

\bibitem{ji2023knowledge}
Z.~Ji, C.~Meng, Y.~Zhang, Y.~Pang, and X.~Li, ``Knowledge-aided momentum
  contrastive learning for remote-sensing image text retrieval,'' {\em IEEE
  Transactions on Geoscience and Remote Sensing}, vol.~61, pp.~1--13, 2023.

\bibitem{wang2022multi}
Y.~Wang, J.~Ma, M.~Li, X.~Tang, X.~Han, and L.~Jiao, ``Multi-scale interactive
  transformer for remote sensing cross-modal image-text retrieval,'' in {\em
  IGARSS 2022-2022 IEEE International Geoscience and Remote Sensing Symposium},
  pp.~839--842, IEEE, 2022.

\bibitem{yuan2023parameter}
Y.~Yuan, Y.~Zhan, and Z.~Xiong, ``Parameter-efficient transfer learning for
  remote sensing image-text retrieval,'' {\em IEEE Transactions on Geoscience
  and Remote Sensing}, 2023.

\bibitem{chen2024integrating}
Y.~Chen, J.~Huang, S.~Xiong, and X.~Lu, ``Integrating multisubspace joint
  learning with multilevel guidance for cross-modal retrieval of remote sensing
  images,'' {\em IEEE Transactions on Geoscience and Remote Sensing}, vol.~62,
  pp.~1--17, 2024.

\bibitem{devlin2018bert}
J.~Devlin, M.-W. Chang, K.~Lee, and K.~Toutanova, ``Bert: Pre-training of deep
  bidirectional transformers for language understanding,'' in {\em North
  American Chapter of the Association for Computational Linguistics}, 2019.

\bibitem{he2020momentum}
K.~He, H.~Fan, Y.~Wu, S.~Xie, and R.~Girshick, ``Momentum contrast for
  unsupervised visual representation learning,'' in {\em Proceedings of the
  IEEE/CVF Conference on Computer Vision and Pattern Recognition},
  pp.~9729--9738, 2020.

\end{thebibliography}
\bibliographystyle{ieeetr}

\end{document}